\documentclass[conference]{IEEEtran}
\IEEEoverridecommandlockouts
\usepackage{cite}
\usepackage{multirow}
\usepackage{amsmath,amssymb,amsfonts}
\usepackage{algorithmic}
\usepackage{graphicx}
\usepackage{textcomp}
\def\BibTeX{{\rm B\kern-.05em{\sc i\kern-.025em b}\kern-.08em
    T\kern-.1667em\lower.7ex\hbox{E}\kern-.125emX}}

\usepackage{fancyhdr}
\begin{document}

\title{Super-pixel cloud detection using \\Hierarchical Fusion CNN\\
{\footnotesize}
\thanks{$^*$ Corresponding author}
\thanks{$^1$ GF-1 is a remote sensing satellite used on earth observation in high resolution.}
}

\author{\IEEEauthorblockN{Han Liu$^\dagger$, Dan Zeng$^\dagger$$^*$ and Qi Tian$^\ddagger$, \emph{Fellow, IEEE}}
\IEEEauthorblockA{\textit{$^\dagger$Key laboratory of Specialty Fiber Optics and Optical Access Networks,} \\
\textit{Joint International Research Laboratory of Specialty Fiber Optics and Advanced Communication,} \\
\textit{Shanghai Institute of  Advanced Communication and Data Science,}\\
\textit{Shanghai University.} \\
Shanghai, China \\
liuhan0122@shu.edu.cn; dzeng@shu.edu.cn}

\IEEEauthorblockA{\textit{$^\ddagger$Department of Computer Science,} \\
\textit{The University of Texas at San Antonio.}\\
San Antonio, United States \\
qi.tian@utsa.edu}
}

\maketitle
\thispagestyle{fancy}

\begin{abstract}
Cloud detection plays a very important role in the process of remote sensing images. This paper designs a super-pixel level cloud detection method based on convolutional neural network (CNN) and deep forest. Firstly, remote sensing images are segmented into super-pixels through the combination of SLIC and SEEDS. Structured forests is carried out to compute edge probability of each pixel, based on which super-pixels are segmented more precisely. Segmented super-pixels compose a super-pixel level remote sensing database. Though cloud detection is essentially a binary classification problem, our database is labeled into four categories: thick cloud, cirrus cloud, building and other culture, to improve the generalization ability of our proposed models. Secondly, super-pixel level database is used to train our cloud detection models based on CNN and deep forest. Considering super-pixel level remote sensing images contain less semantic information compared with general object classification database, we propose a Hierarchical Fusion CNN (HFCNN). It takes full advantage of low-level features like color and texture information and is more applicable to cloud detection task. In test phase, every super-pixel in remote sensing images is classified by our proposed models and then combined to recover final binary mask by our proposed distance metric, which is used to determine ambiguous super-pixels. Experimental results show that, compared with conventional methods, HFCNN can achieve better precision and recall.
\end{abstract}


\section{Introduction}
With the development of remote sensing technology, remote sensing images are widely used in various fields, such as land-use program, meteorological monitoring, geological exploration and so on. But sensors are often influenced by atmosphere density and cloud changing. Clouds in remote sensing images have some specific characteristics including brightness, color, texture, shape, etc. Some buildings reflecting sun light look like clouds in high resolution remote sensing images like GF-1$^1$. Many remote sensing images are troubled by cloud occlusion, which causes more difficult observations on remote sensing images and inaccurate analysis results. So cloud detection on remote sensing images becomes more meaningful if we can locate cloud region or even recover original culture on the surface\cite{b1}.

Humans generally identify clouds roughly by analysing local and global features of remote sensing images. Most previous methods perform pixel level cloud detection, however only using information of pixels may be unreliable. Super-pixels merge pixels with similar visual features into pixel blocks. Super-pixel level cloud detection methods using super-pixels as basic units can also achieve excellent results compared with pixel level methods. In this paper, an automatic super-pixel level cloud detection method for remote sensing images is proposed. This paper focuses on cloud detection task on super-pixel level GF-1 remote sensing images using CNN and deep forest\cite{b2}. The steps include super-pixel segmentation, building super-pixel level remote sensing database, training HFCNN and deep forest, refinement by distance metric using predictions and features from two models.

The contributions of this work are four-fold:
\begin{itemize}
\item Considering remote sensing super-pixels have less semantic information, Hierarchical Fusion CNN (HFCNN) is proposed to make the best of low-level features such as color, texture and shape.
\item Structured forests \cite{b3} is implemented for edge detection on remote sensing images, combined with SLIC and SEEDS for more accurate super-pixel segmentation.
\item Based on predictions and features from CNN and deep forest, we propose super-pixel level distance metric to further refine ambiguous super-pixels which might be cloud utilizing super-pixels surrounding them.
\item A deep forest model is trained for cloud detection task, which also results in exciting performance. With the combination of HFCNN, we achieve the best super-pixel level precision and recall.
\end{itemize}

The remainder of this paper is organized as follows. Section \ref{RW} introduces the existing work. Section \ref{DNN} discusses the proposed remote sensing super-pixel cloud detection in details, including how to label super-pixels in large remote sensing images, how to train HFCNN, and how to further refine ambiguous super-pixels using our proposed distance metric. Finally, experimental results and analysis are presented in section \ref{ER}, and section \ref{CAFW} concludes the paper.

\section{Related work}\label{RW}
Remote sensing cloud detection methods can be roughly divided into two categories: threshold-based methods and machine-learning-based methods. Most cloud detection methods extract features pixel by pixel, followed by setting a threshold or learning a binary classifier to determine whether a pixel is cloud or not \cite{b1}. The key to threshold-based methods is how to choose the optimum threshold to distinguish foreground cloud from background surface. Early methods with fixed threshold have failed to meet the increasing precision requirements. Paying attention to differences between cloud and surface features, more and more dynamic and adaptive threshold methods are proposed. Spatially and temporally varying thresholds were incorporated into cloud detection process using two different channels images \cite{b4}. Zhang and Xiao \cite{b5} proposed an automatic and efficient cloud detection algorithm based on observations and statistics on remote sensing images. They improved Otsu, a classic global threshold method and progressively refined detection results.

\begin{figure}[htbp]
\centerline{\includegraphics[width=6cm]{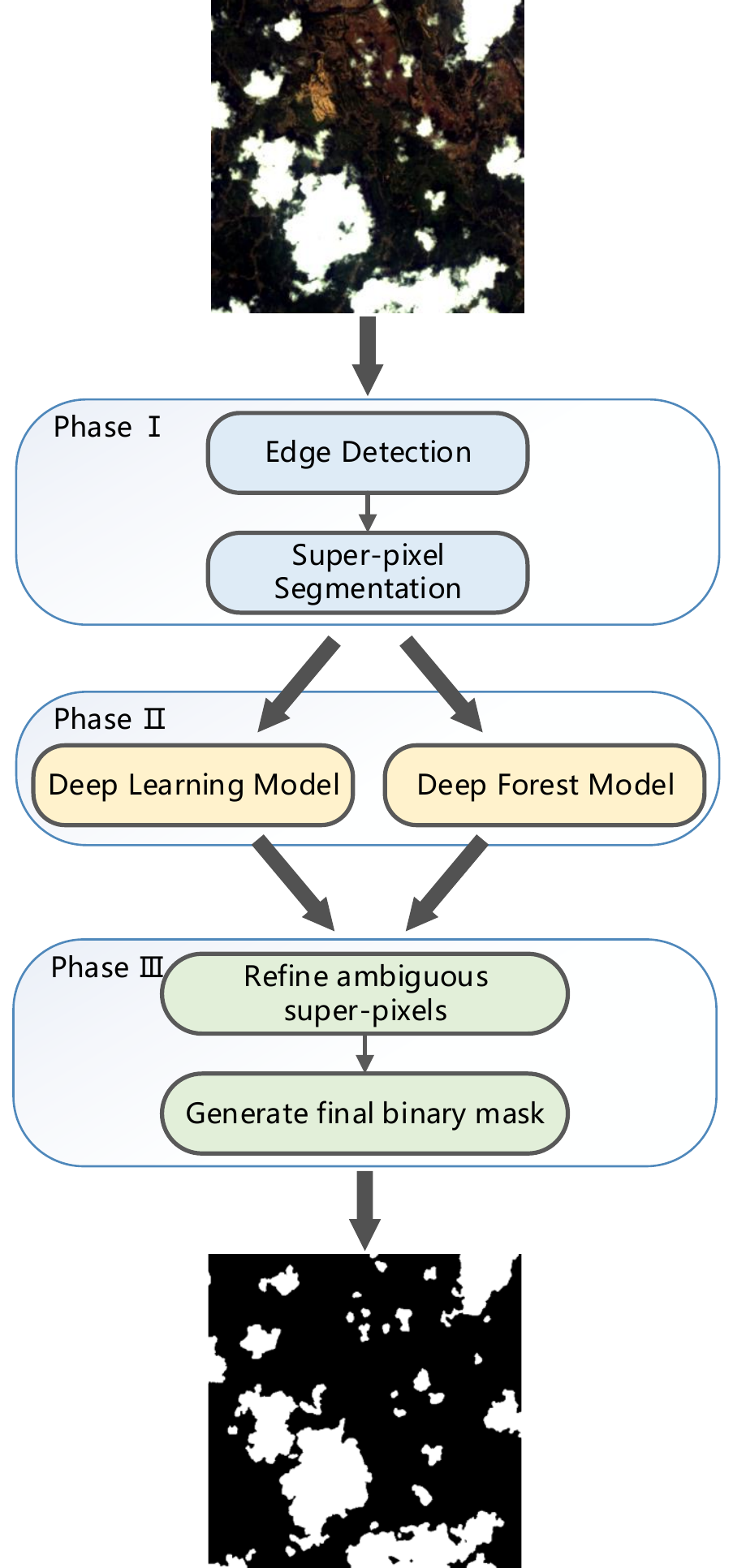}}
\caption{Overall architecture.}
\label{fig1}
\end{figure}

In recent years, machine learning has achieved excellent performance in many fields, e.g. image classification, semantic segmentation and edge detection. Generally, training a machine learning model requires a large scale of samples. However, for remote sensing cloud detection, sensors are easily influenced by atmosphere environment, and there is no public large scale database, which makes cloud detection difficult when using machine learning. Visa \cite{b6} implemented neural network models on cloud detection based on texture segmentation, they didn't mention how they label training data in spite of good performance. It would be time-consuming, laborious and prone to errors if remote sensing images are labeled manually pixel by pixel. Instead, we use super-pixel level remote sensing images, it would be much easier to label super-pixels than a whole remote sensing image pixel by pixel. Xie et al. \cite{b1} improved conventional super-pixel segmentation method SLIC, and trained a model using CNN. We also implement super-pixel level cloud detection, but using different segmentation method. Hu \cite{b7} classified remote sensing images into three saliency maps using random forest classifier, and a threshold is used on final saliency map to determine whether images contain cloud or not. Tan \cite{b8} combined several statistical features of one super-pixel resulting good performance with SVM model. Yuan and Hu \cite{b9} also proposed object-based cloud detection using Bag-of-Words (BoW) feature representation. Low-level features like SIFT are extracted and map into BoW vectors, then SVM is used to make a classification. Instead of handcrafted features followed by SVM, we adopt an end-to-end CNN model which combines feature extractor and classifier.

Deep forest \cite{b2} was first proposed by Zhou and Feng in 2017. They think deep forest can be an alternative of CNN. Many computer vision tasks have applied deep forest \cite{b10,b11,b12}. In a way, deep forest can achieve comparable accuracy to CNN in some tasks. There is no contributions on cloud detection using deep forest by far, our work is a new attempt at deep forest application.

\section{Convolutional Neural Network Based Super-pixel Cloud Detection}\label{DNN}
\begin{figure}[htbp]
\begin{minipage}{0.45\linewidth}
  \centerline{\includegraphics[width=4cm]{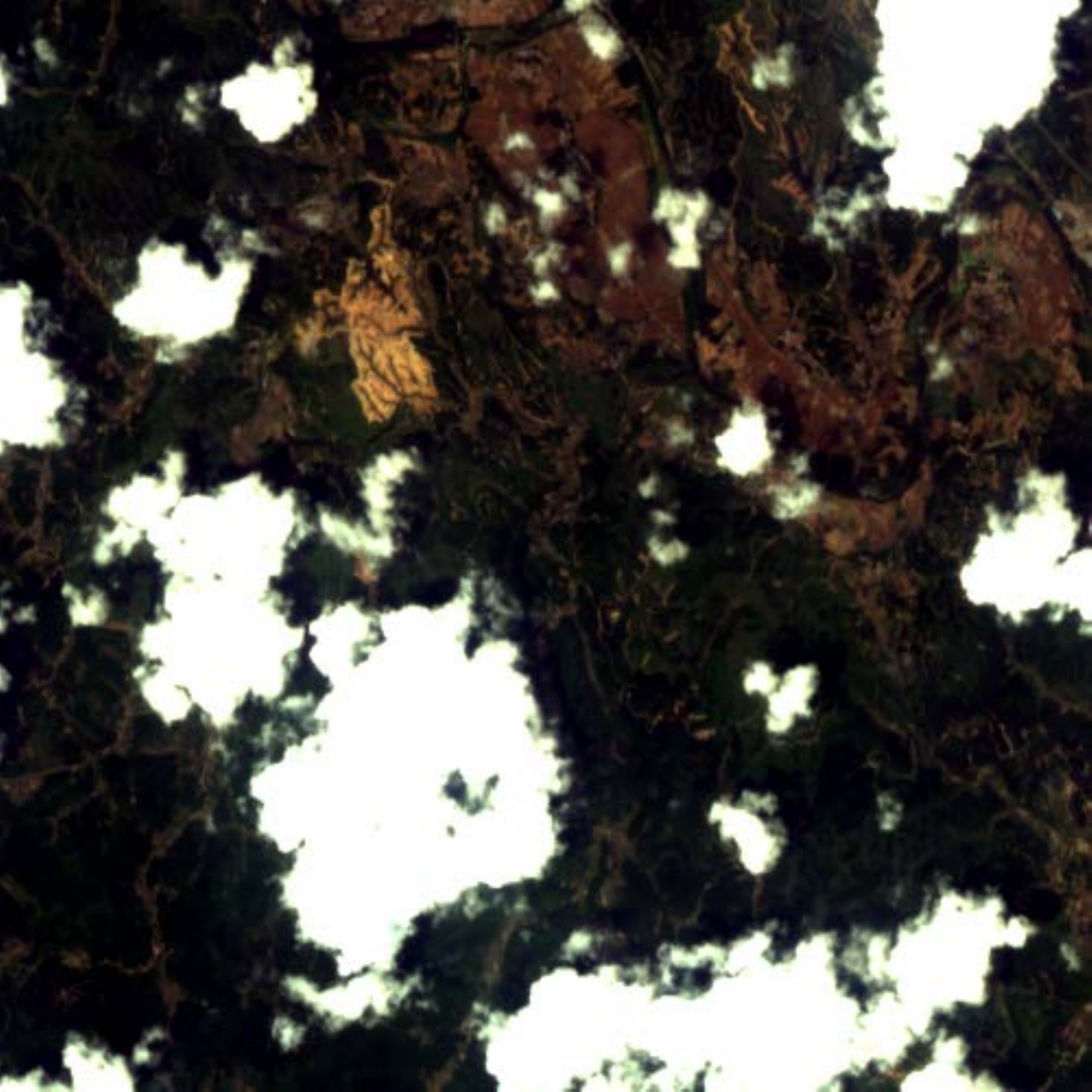}}
  \centerline{(a)}
\end{minipage}
\hfill
\begin{minipage}{0.45\linewidth}
  \centerline{\includegraphics[width=4cm]{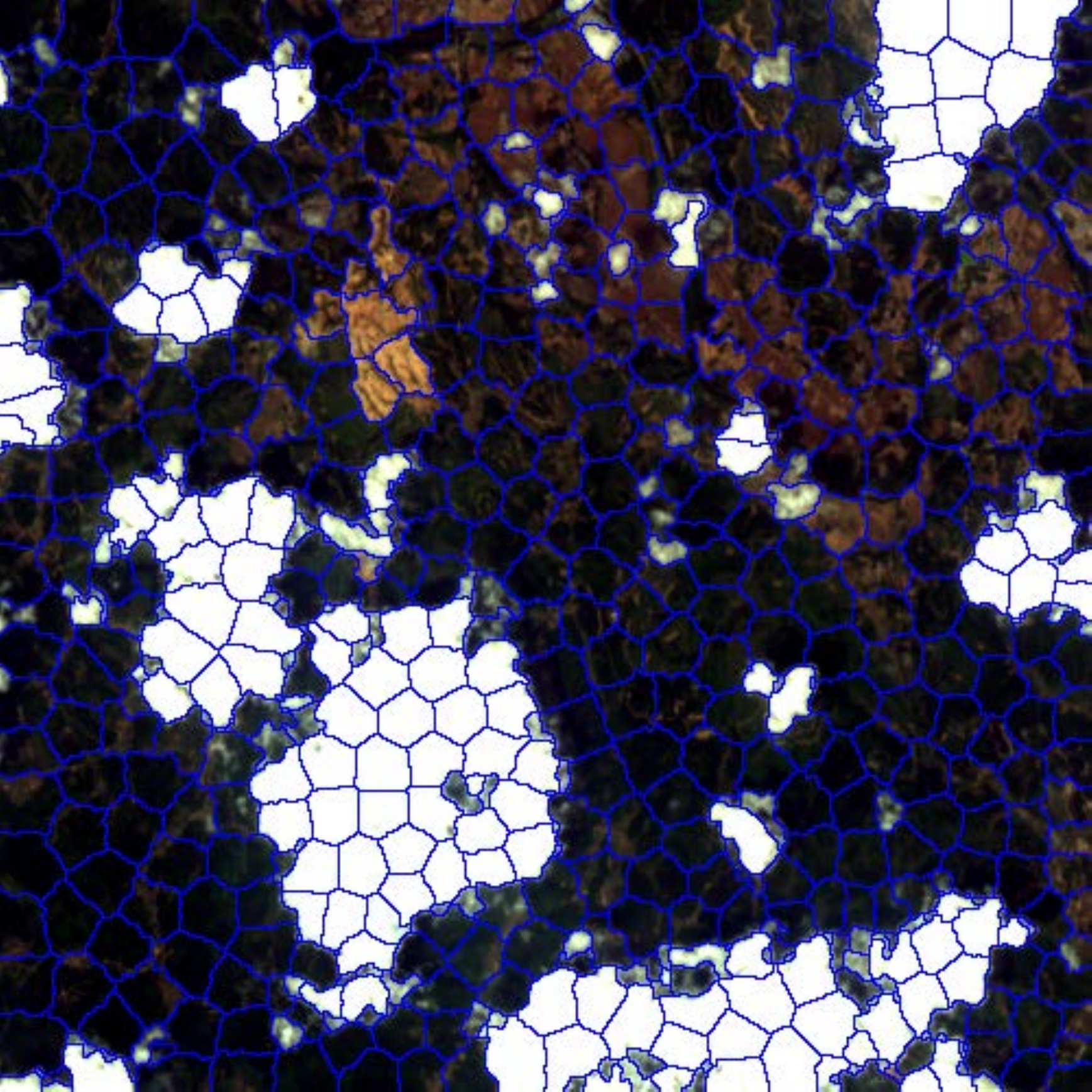}}
  \centerline{(b)}
\end{minipage}
\caption{Illustration of super-pixel segmentation. (a) Original remote sensing image. (b) Remote sensing image after super-pixel segmentation.}
\label{fig2}
\end{figure}
The overall architecture is illustrated in Fig.\ref{fig1}. The whole process can be roughly divided into three phases. In section \ref{AA}, phase one, remote sensing images are segmented into super-pixels, which compose a super-pixel level remote sensing database. In section \ref{BB}, phase two, the proposed Hierarchical Fusion CNN (HFCNN) and a deep forest model are trained. Two models output their predictions and features of super-pixels separately. We combine two kinds of features from both HFCNN and deep forest to get more accurate results and further refine ambiguous super-pixels by our proposed super-pixel level distance metric in phase three, section \ref{CC}.
\subsection{Build Super-pixel cloud database}\label{AA}
CNN is driven by large scale database. All computer vision tasks can hardly get significant results without large amount of training data. The same goes for deep forest. In this case, a great quantity of super-pixel level remote sensing images are needed to ensure robustness of our models.

Structured forests\cite{b3} is applied to implement edge detection on remote sensing images. Structured forests is a very fast edge detector that achieves excellent accuracy. The super-pixel segmentation algorithm which detects sticky edge adhesive super-pixels is affiliate with structured forests. With structured forests edge detection, segmented super-pixels snap to edges, resulting in higher quality boundaries. Super-pixels are segmented using an iterative approach motivated by both SLIC and SEEDS. Fig.\ref{fig2} (a) is the original remote sensing image, and the super-pixel segmentation result is shown in Fig.\ref{fig2} (b).

\begin{figure}[htbp]
\centerline{\includegraphics[width=8cm]{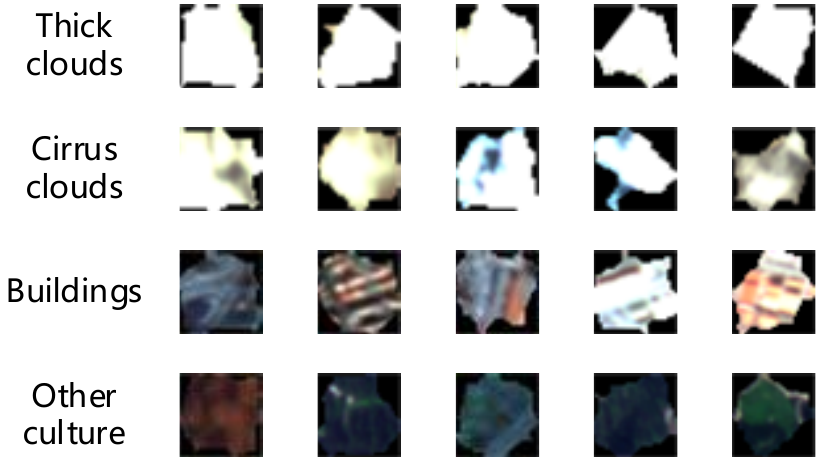}}
\caption{Illustration of super-pixel level remote sensing cloud detection database.}
\label{fig3}
\end{figure}

To build a super-pixel level remote sensing database, we need to label the segmented super-pixels as shown in Fig.\ref{fig2} (b) one by one. The method we label super-pixels is both accurate and convenient compared with \cite{b6}. The whole database has 2800 $32 \times 32$ super-pixel level images, some typical examples are shown in Fig.\ref{fig3}. We classify these super-pixels into four categories: thick cloud, cirrus cloud, building and other culture, in spite cloud detection is a binary classification task. As clouds located at the boarder of thick cloud are always blurry, uncertain and irregular, we add a ``cirrus cloud'' category. Similarly, buildings might look like thick cloud when they reflect sun light, like the super-pixel at third row and the fourth column in Fig.\ref{fig3}, so we add a ``building'' category. In this case, our model has better generalization ability and robustness, and we can process these ambiguous super-pixels after prediction from HFCNN and deep forest.
\subsection{Hierarchical Fusion CNN and deep forest}\label{BB}
After building the super-pixel level remote sensing database, CNN and deep forest are trained separately. Features and predictions from two models are used for accurate classification. Compared with conventional methods extracting complicated handcrafted features and sophisticated classifiers, CNN and deep forest can automatically learn local details and semantic features. We find three characteristics from our labeled database: (1) Super-pixels have less semantic information. (2) Super-pixels have simple texture feature. (3) Super-pixels have small size.

Considering three characteristics mentioned above, we choose the simple CIFAR-10 Fast Model as backbone, and modify it. CIFAR-10 \cite{b13} is a database used for general object classification, and CIFAR-10 Fast Model is a simple network used to train CIFAR-10 database, which is included in Caffe examples. We call it CIFAR-10 network for short. Generally speaking, fine-tuning existing model on our own database can achieve better result than training from scratch. But there is vast bias between CIFAR-10 database and our own database. Images in CIFAR-10 database contain more semantic information, e.g. car, bird, car. But our super-pixel level remote sensing images have less semantic information but more basic color and texture information. For example, thick cloud super-pixels have obvious color feature without texture, while most super-pixels in building category have clear texture feature. So we modify original CIFAR-10 network and propose Hierarchical Fusion CNN (HFCNN) to make the best of low-level features and make it more suitable for cloud detection task.

\begin{figure}[htbp]
\centerline{\includegraphics[width=6cm]{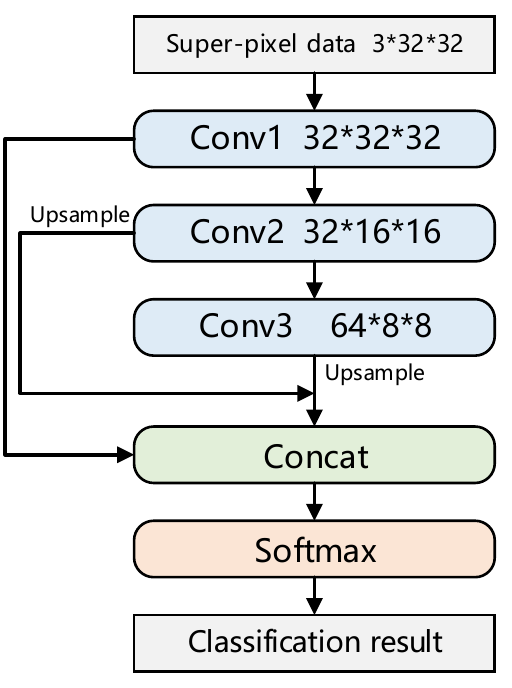}}
\caption{Architecture of HFCNN.}
\label{fig4}
\end{figure}

Network architecture is shown in Fig.\ref{fig4}. Our model is simple, intuitionistic and doesn't need to extract complicated feature compared with\cite{b8}, which combined several features of super-pixels and trained with SVM. In the first few layers, e.g. Convolution layer 1, Convolution layer 2, CNN can extract low-level features like texture and color. In deep layers, e.g. Convolution layer 3, CNN can extract abstract features which are important for tasks like object classification. As super-pixels in our database have less semantic information, we should pay more attention on features extracted from first few layers. Inspired by FPN\cite{b14}, we propose HFCNN. HFCNN up-samples feature maps from Convolution layer 2 and Convolution layer 3, then concatenate them with feature map from Convolution layer 1 channel-wisely. This fused feature is sent into Softmax layer for classification. In this case, HFCNN can take full advantages of low-level features.

As for deep forest, we use the network proposed in \cite{b2} to train our model directly. Deep forest needs less hyper parameter or training data than CNN, which makes it more efficient. Deep forest is composed of completely-random tree forests and random forests. Each forest consists of 500 decision trees. Sliding windows is first operated on input images, then features will be extracted in completely-random tree forests and random forests. This step is called multi-grained scanning. Then features are sent into cascade forest for feature extracting and classification layer by layer.
\subsection{Refinement and final binary mask}\label{CC}
Unlike methods utilizing information of multi-channel and spectrum to detect cloud, we only extract features from RGB images with HFCNN and deep forest. To achieve more accurate results, further refinement is performed to achieve better precision and recall. In this step, a super-pixel level distance metric is proposed for measuring similarity between super-pixels and determining categories of ambiguous super-pixels.

Some segmented super-pixels might be too small to get accurate predictions from models, we need to remove those super-pixels whose area are smaller than a certain threshold. In experiment, the threshold is set to 10.

GF-1 remote sensing images have four channels in total, first three are RGB color channels, and the last one is near-infrared channel. Actually, near-infrared sensor is sensitive to cloud. Its response to cloud is always larger than culture. So we can set a threshold on data from near-infrared channel to further confirm whether super-pixels are cloud or not, such as removing false positive super-pixels like white buildings. So far, super-pixels predicted as thick cloud are definitely clouds.

\begin{figure}[htbp]
\begin{minipage}{0.3\linewidth}
  \centerline{\includegraphics[height=1.5cm]{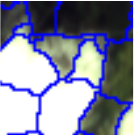}}
  \centerline{(a)}
\end{minipage}
\hfill
\begin{minipage}{0.3\linewidth}
  \centerline{\includegraphics[height=1.5cm]{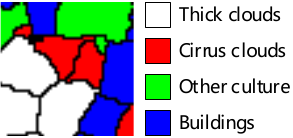}}
  \centerline{(b)}
\end{minipage}
\hfill
\begin{minipage}{0.3\linewidth}
  \centerline{\includegraphics[height=1.5cm]{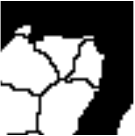}}
  \centerline{(c)}
\end{minipage}

\caption{(a) Remote sensing image after super-pixel segmentation. (b) Results of classification after HFCNN. (c) Result after refinement.}
\label{fig5}
\end{figure}

HFCNN and deep forest trained in section \ref{BB} can output not only predictions of super-pixels, but also high-dimensional features of them. When computing distance between classified super-pixels, we tried using features from HFCNN, deep forest and concatenation of both to compute distance metric between super-pixels, and we found features from the concatenation of HFCNN and deep forest achieve best results. Our proposed distance metric is used to determine categories of super-pixels located at the boarder of thick clouds like cirrus clouds and buildings. Firstly, we get distance matrix in which cosine similarity between each super-pixel and super-pixels surrounding it are computed. It is worth noting that it's meaningless and computational-consuming to compute distances with super-pixels far away. Secondly, our proposed distance metric is used on super-pixels which are predicted as cirrus cloud and building by HFCNN and deep forest. If a cirrus cloud or building super-pixel satisfies the following equation, its category will be changed to thick cloud.

\begin{small}
\begin{equation}
distance(S_{thick},S_{cirrus})<distance(S_{other},S_{cirrus})
\label{eq1}
\end{equation}
\end{small}
\begin{small}
\begin{equation}
distance(S_{thick},S_{cirrus})\!=\!\frac{\sum_{1}^{n}(S_{thick}\!\times\! S_{cirrus})}
{\sqrt{\sum_{1}^{n}S_{thick}^{2}}\!\times\! \sqrt{\sum_{1}^{n}S_{cirrus}^{2}}}
\label{eq2}
\end{equation}
\end{small}
\begin{small}
\begin{equation}
distance(S_{other},S_{cirrus})\!=\!\frac{\sum_{1}^{n}(S_{other}\!\times\! S_{cirrus})}
{\sqrt{\sum_{1}^{n}S_{other}^{2}}\!\times\! \sqrt{\sum_{1}^{n}S_{cirrus}^{2}}}
\label{eq2}
\end{equation}
\end{small}

where $S_{thick}$ represents feature of thick cloud super-pixel, like the white super-pixels shown in Fig.\ref{fig5} (b). $S_{cirrus}$ represents feature of cirrus cloud super-pixel, like the red super-pixels shown in Fig.\ref{fig5} (b). $S_{other}$ represents feature of other categories, like green and blue super-pixels shown in Fig.\ref{fig5} (b). $distance(S_{thick},S_{cirrus})$ represents difference between $S_{thick}$ and $S_{cirrus}$. If $distance(S_{thick},S_{cirrus})<distance(S_{other},S_{cirrus})$, the cirrus cloud is more similar with thick cloud instead of other categories. Fig.\ref{fig5} (c) shows the result after refinement, cirrus clouds are detected accurately.

As we can see in Fig.\ref{fig5} (c), there are gaps between detect cloud super-pixels. Finding boundaries of super-pixels, gaps can be eliminated easily.

\section{Experimental Results}\label{ER}
\begin{figure*}[htbp]
\begin{minipage}{0.16\linewidth}
  \centerline{\includegraphics[width=2.9cm]{11.pdf}}
  \centerline{ }
\end{minipage}
\hfill
\begin{minipage}{0.16\linewidth}
  \centerline{\includegraphics[width=2.9cm]{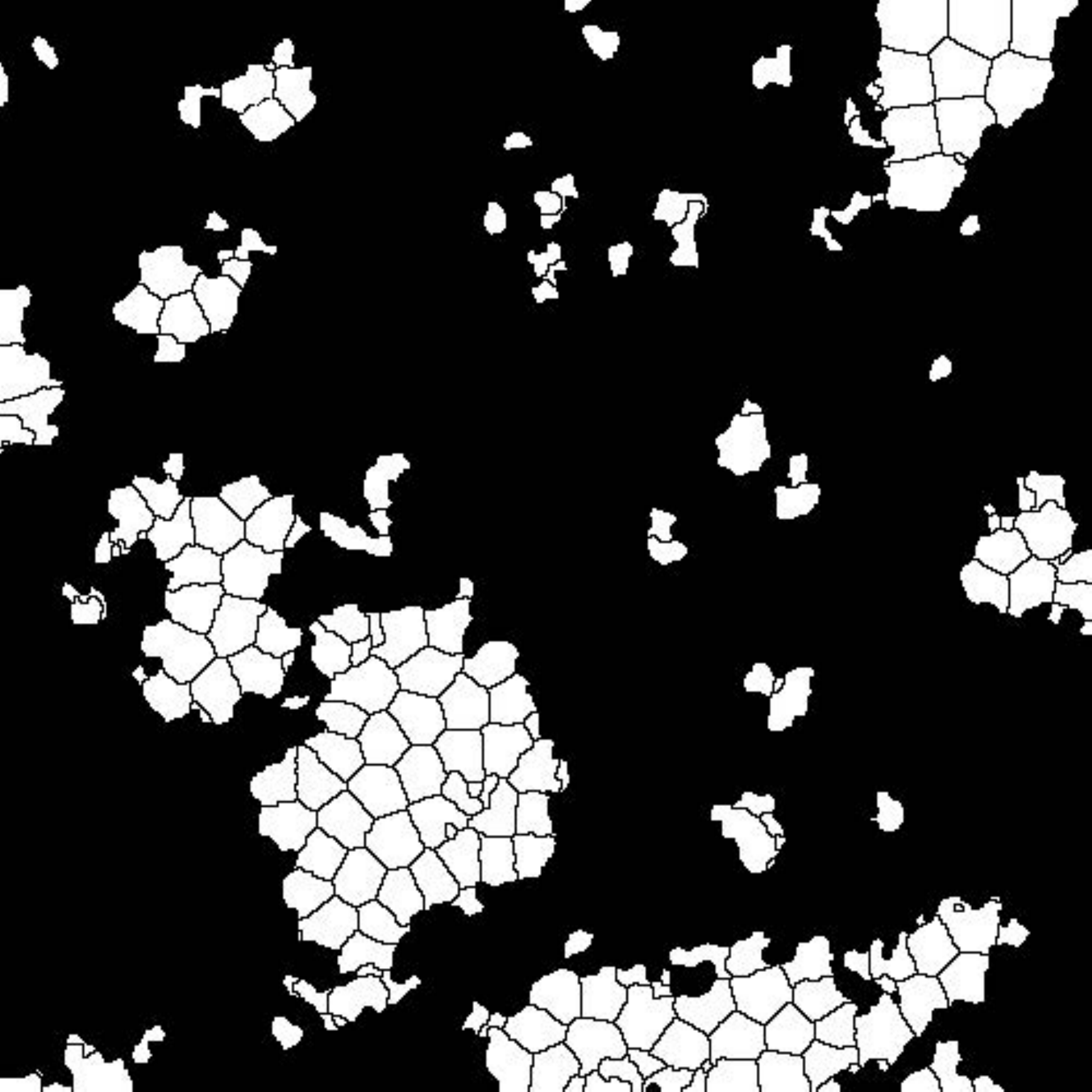}}
  \centerline{ }
\end{minipage}
\hfill
\begin{minipage}{0.16\linewidth}
  \centerline{\includegraphics[width=2.9cm]{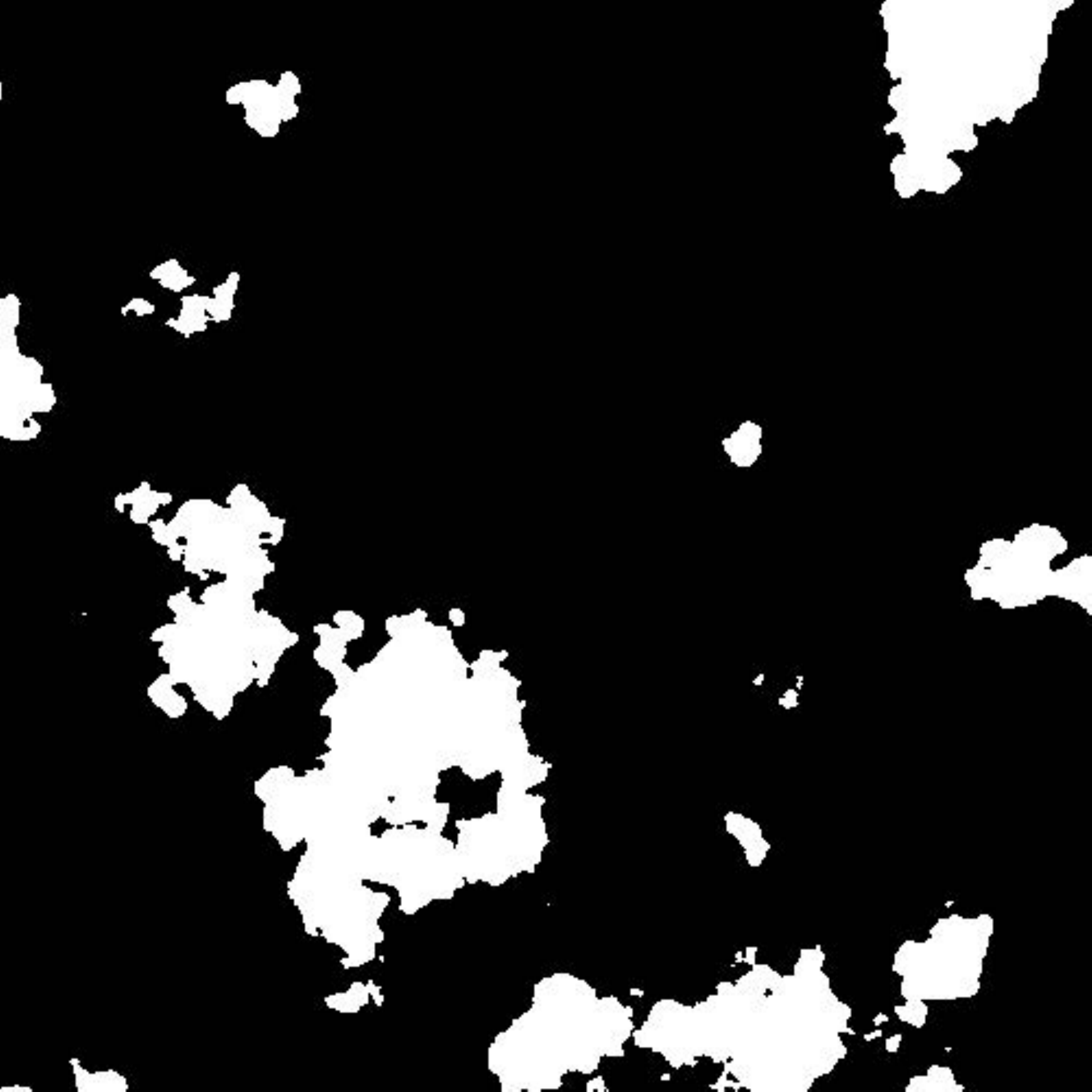}}
  \centerline{ }
\end{minipage}
\hfill
\begin{minipage}{0.16\linewidth}
  \centerline{\includegraphics[width=2.9cm]{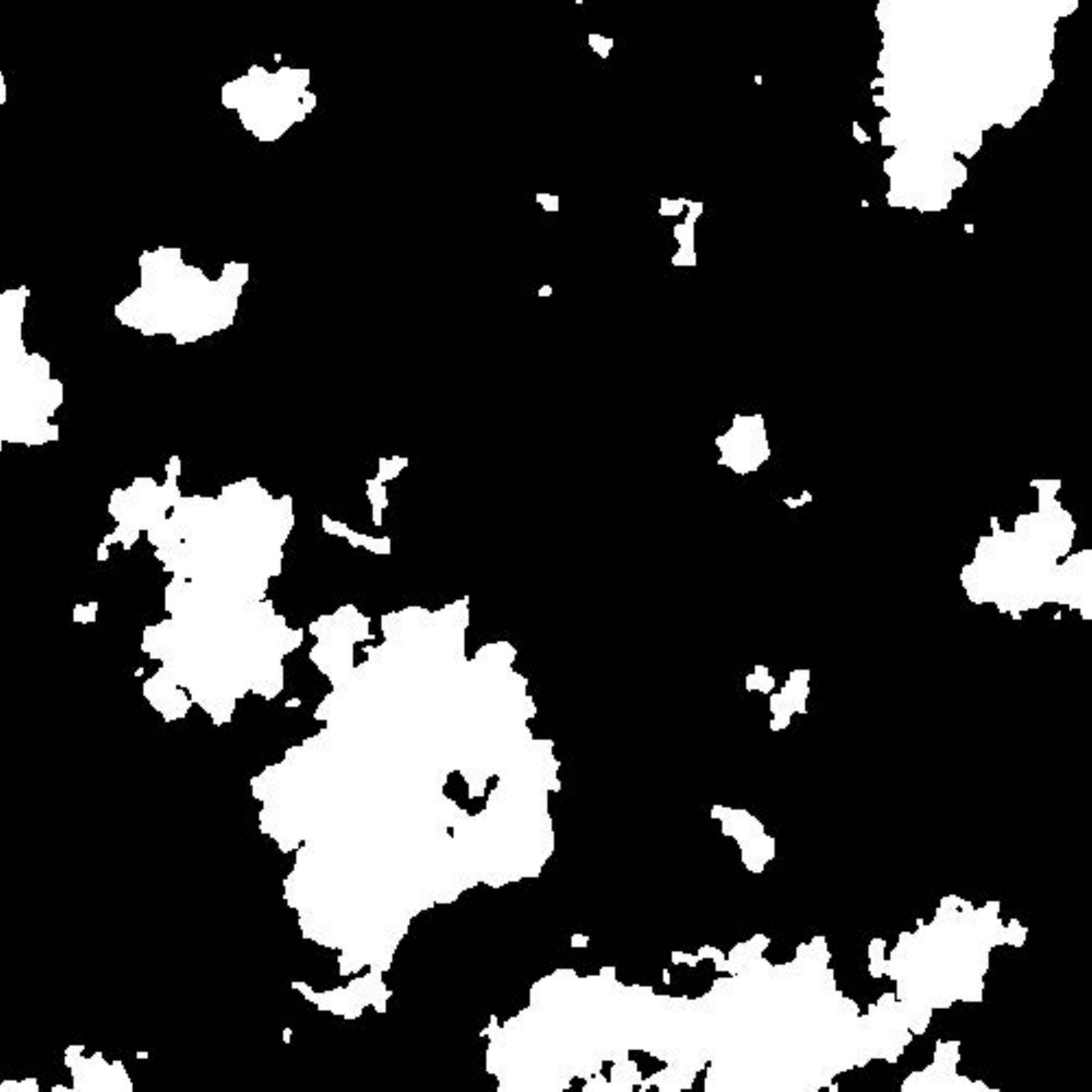}}
  \centerline{ }
\end{minipage}
\hfill
\begin{minipage}{0.16\linewidth}
  \centerline{\includegraphics[width=2.9cm]{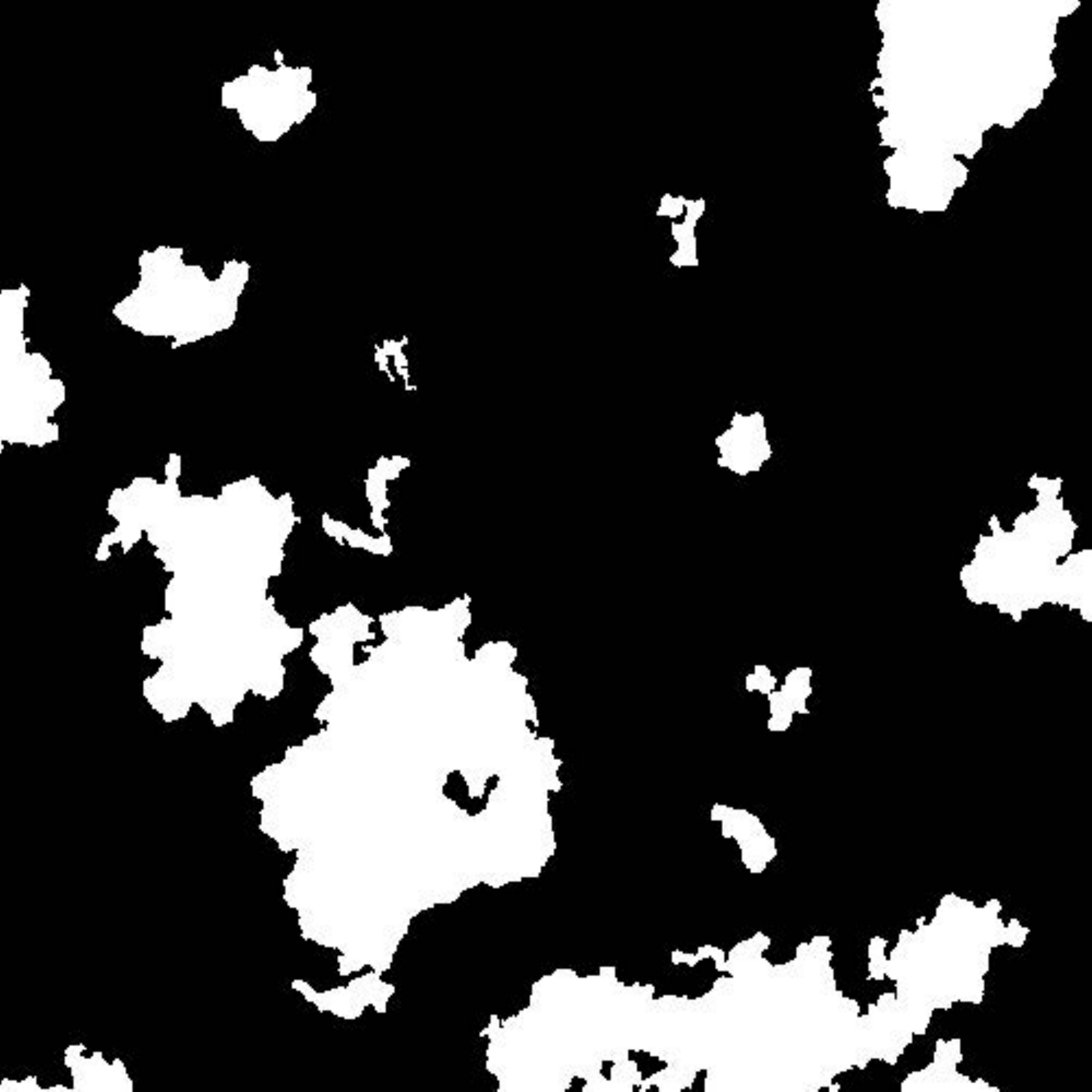}}
  \centerline{ }
\end{minipage}
\hfill
\begin{minipage}{0.16\linewidth}
  \centerline{\includegraphics[width=2.9cm]{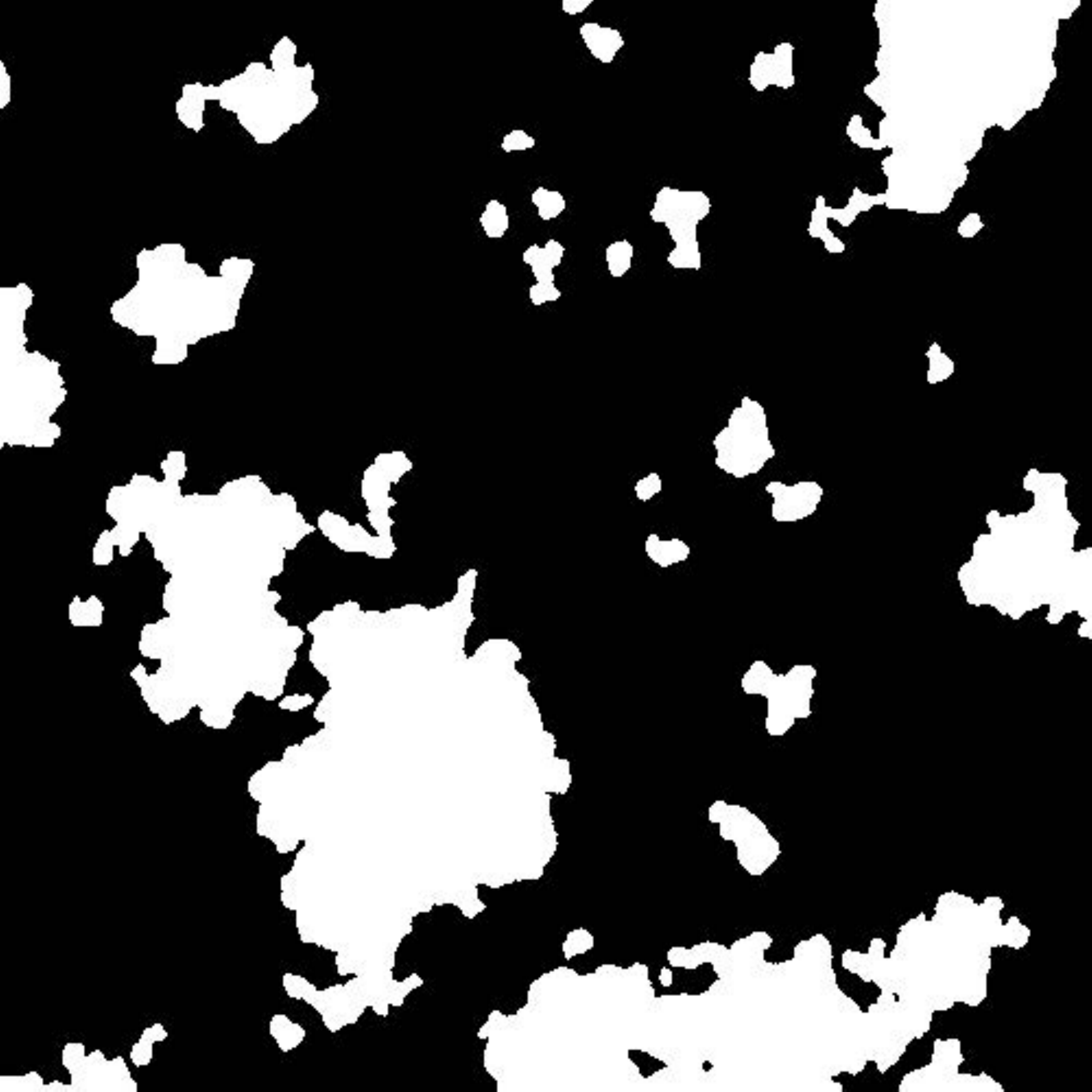}}
  \centerline{ }
\end{minipage}
\hfill
\begin{minipage}{0.16\linewidth}
  \centerline{\includegraphics[width=2.9cm]{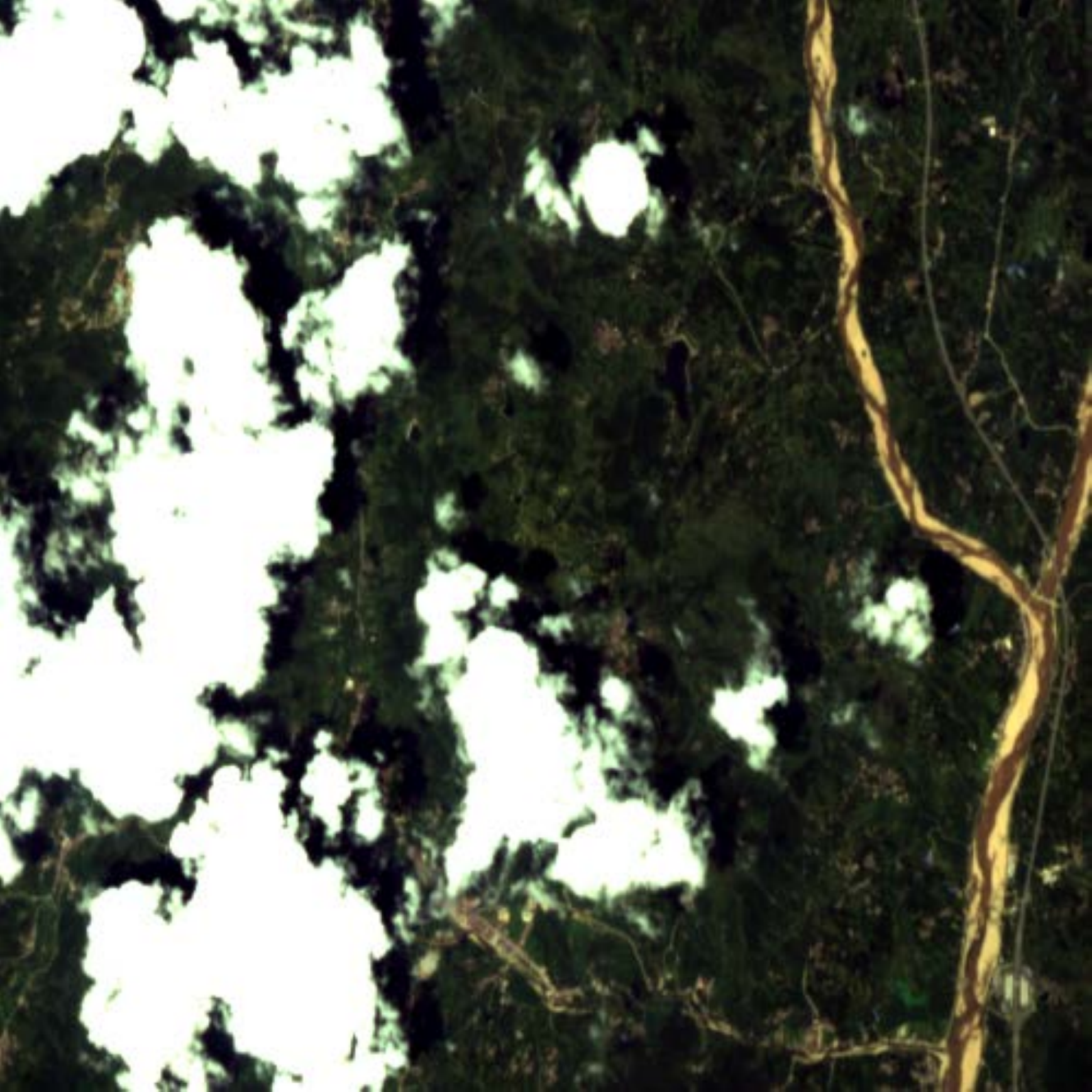}}
  \centerline{ }
\end{minipage}
\hfill
\begin{minipage}{0.16\linewidth}
  \centerline{\includegraphics[width=2.9cm]{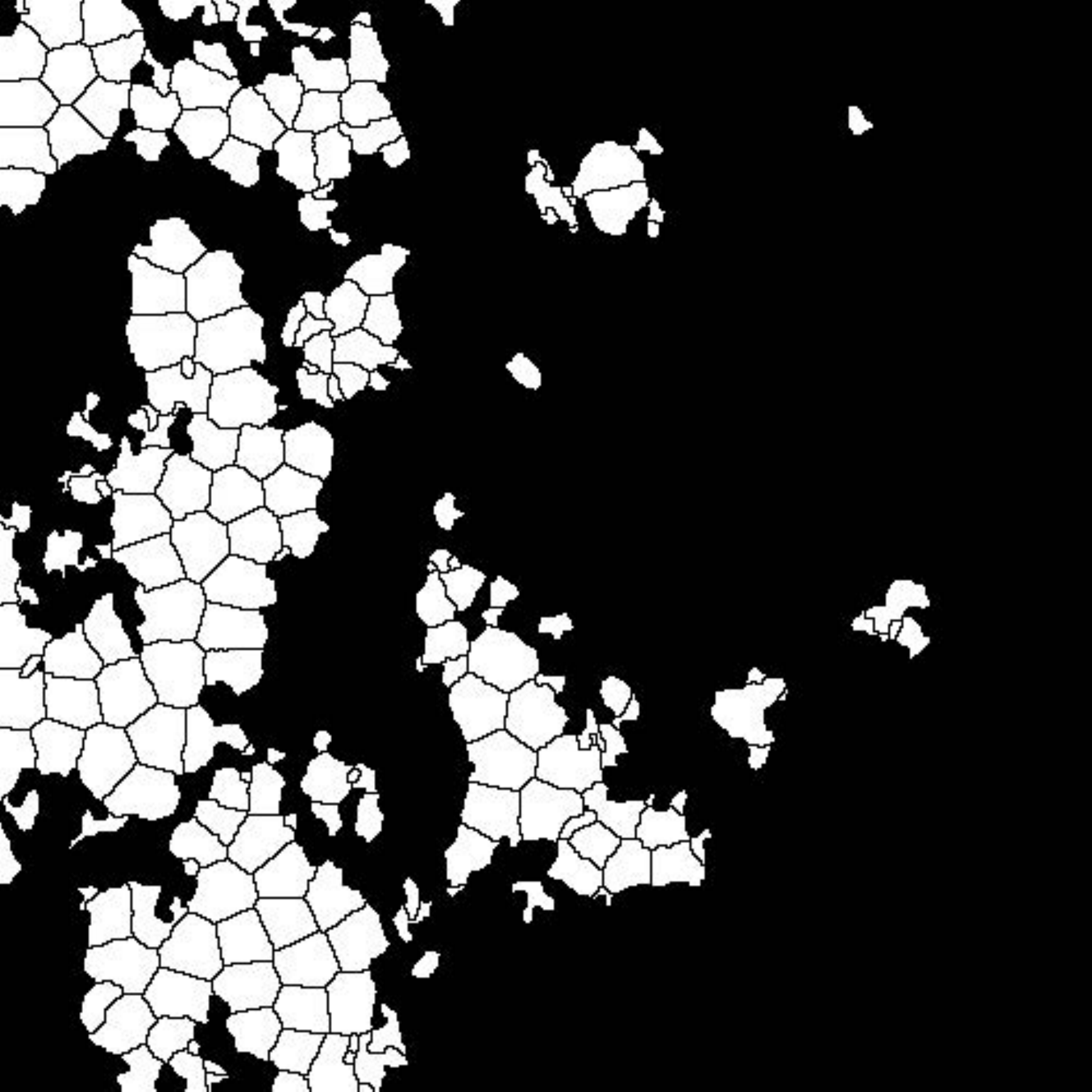}}
  \centerline{ }
\end{minipage}
\hfill
\begin{minipage}{0.16\linewidth}
  \centerline{\includegraphics[width=2.9cm]{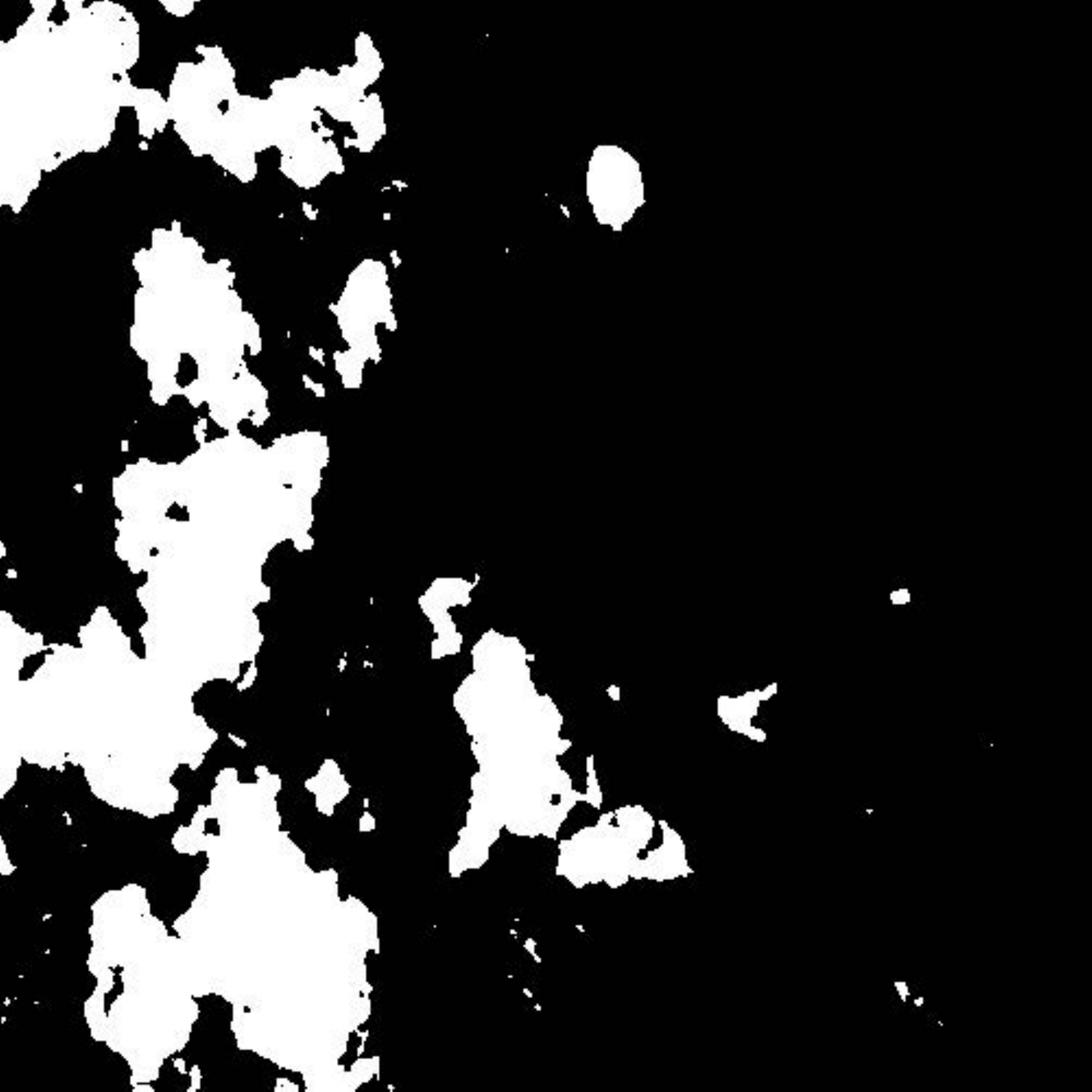}}
  \centerline{ }
\end{minipage}
\hfill
\begin{minipage}{0.16\linewidth}
  \centerline{\includegraphics[width=2.9cm]{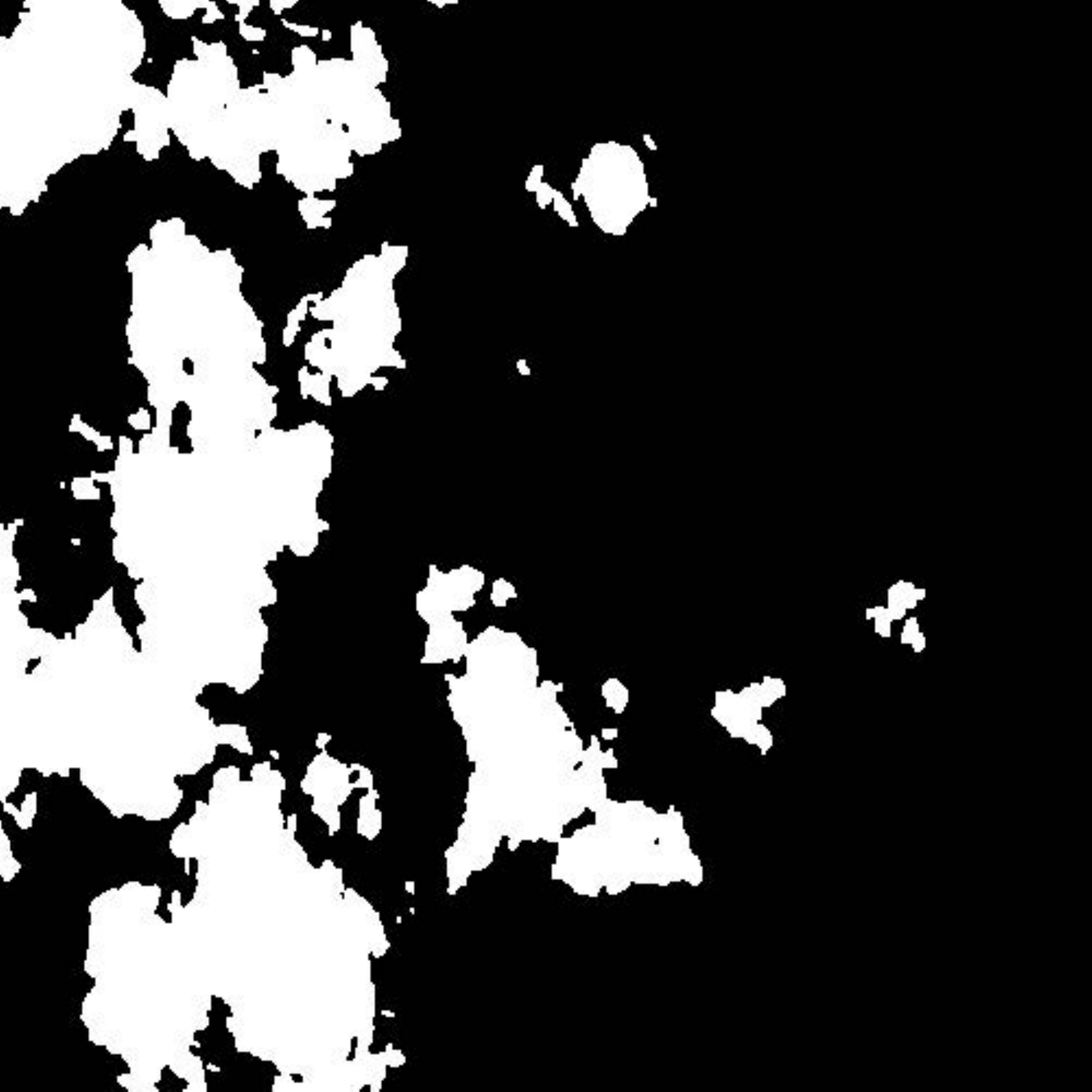}}
  \centerline{ }
\end{minipage}
\hfill
\begin{minipage}{0.16\linewidth}
  \centerline{\includegraphics[width=2.9cm]{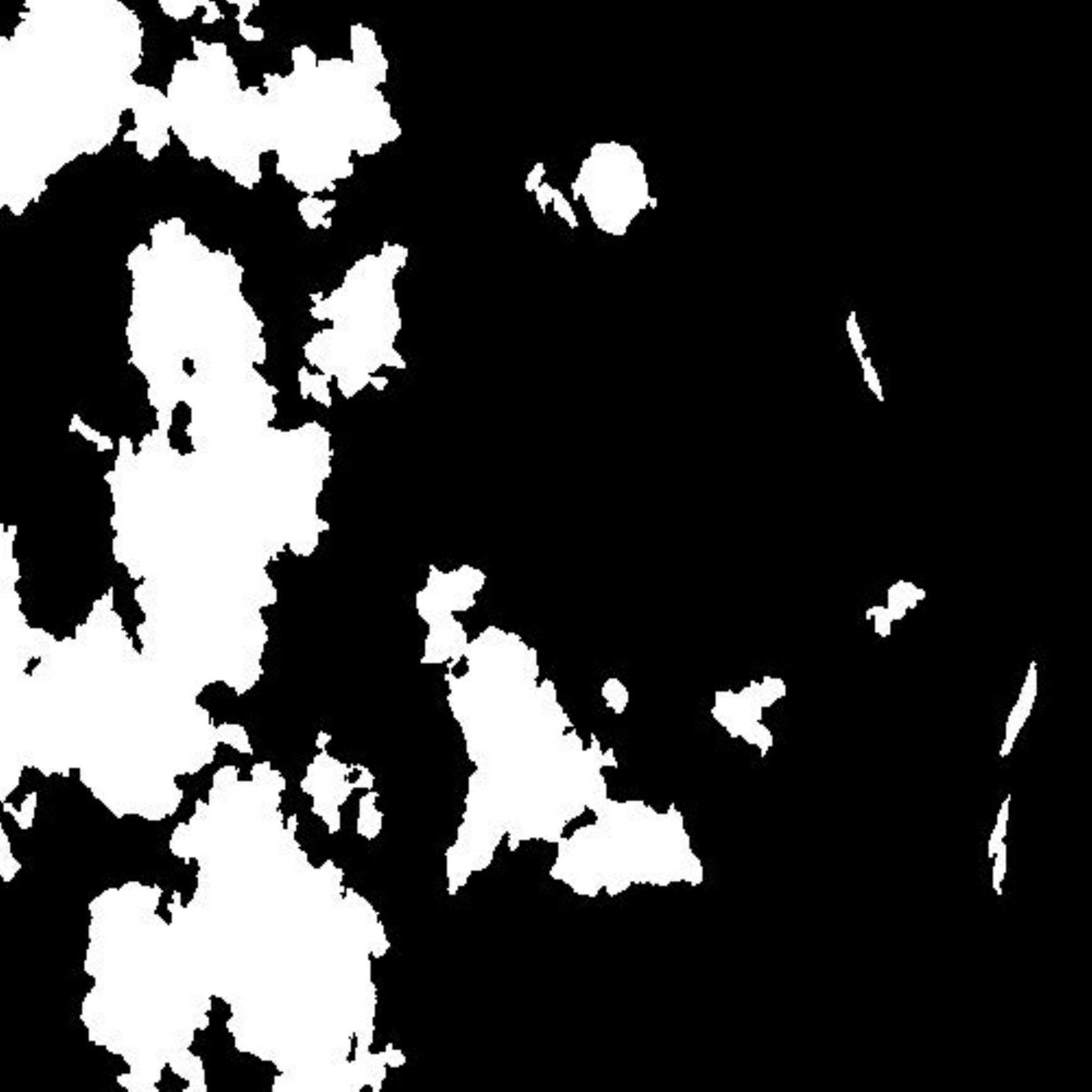}}
  \centerline{ }
\end{minipage}
\hfill
\begin{minipage}{0.16\linewidth}
  \centerline{\includegraphics[width=2.9cm]{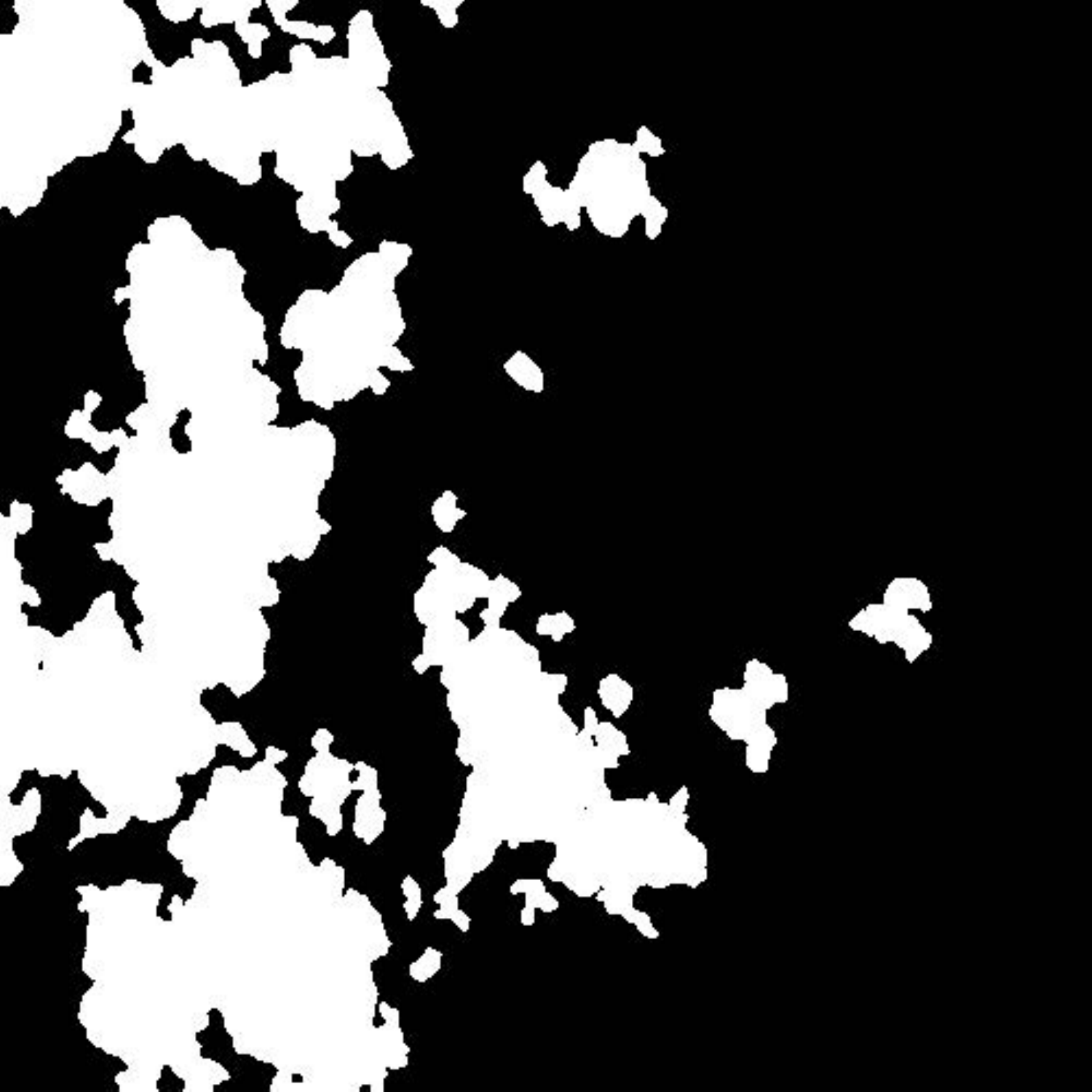}}
  \centerline{ }
\end{minipage}
\hfill
\begin{minipage}{0.16\linewidth}
  \centerline{\includegraphics[width=2.9cm]{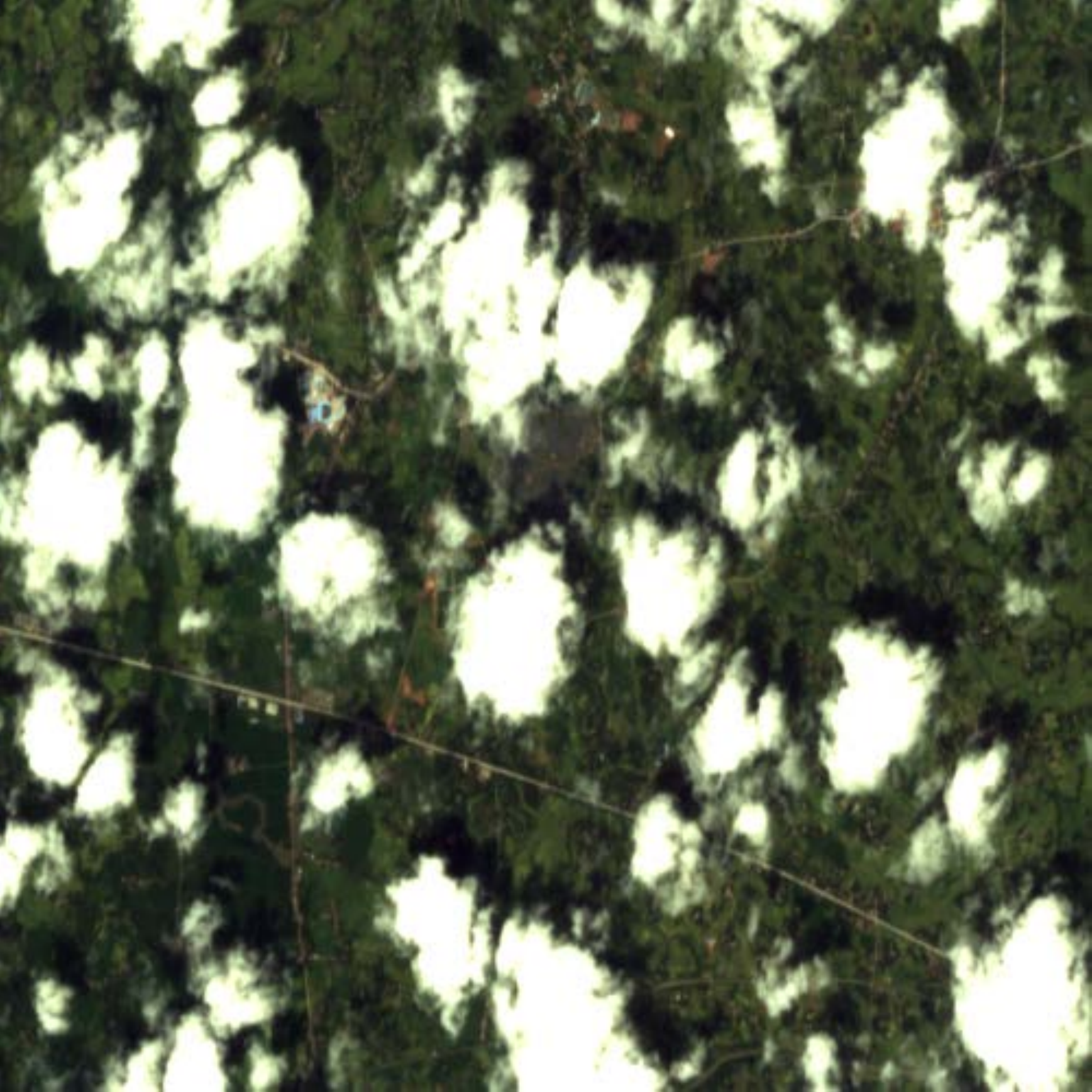}}
  \centerline{ }
\end{minipage}
\hfill
\begin{minipage}{0.16\linewidth}
  \centerline{\includegraphics[width=2.9cm]{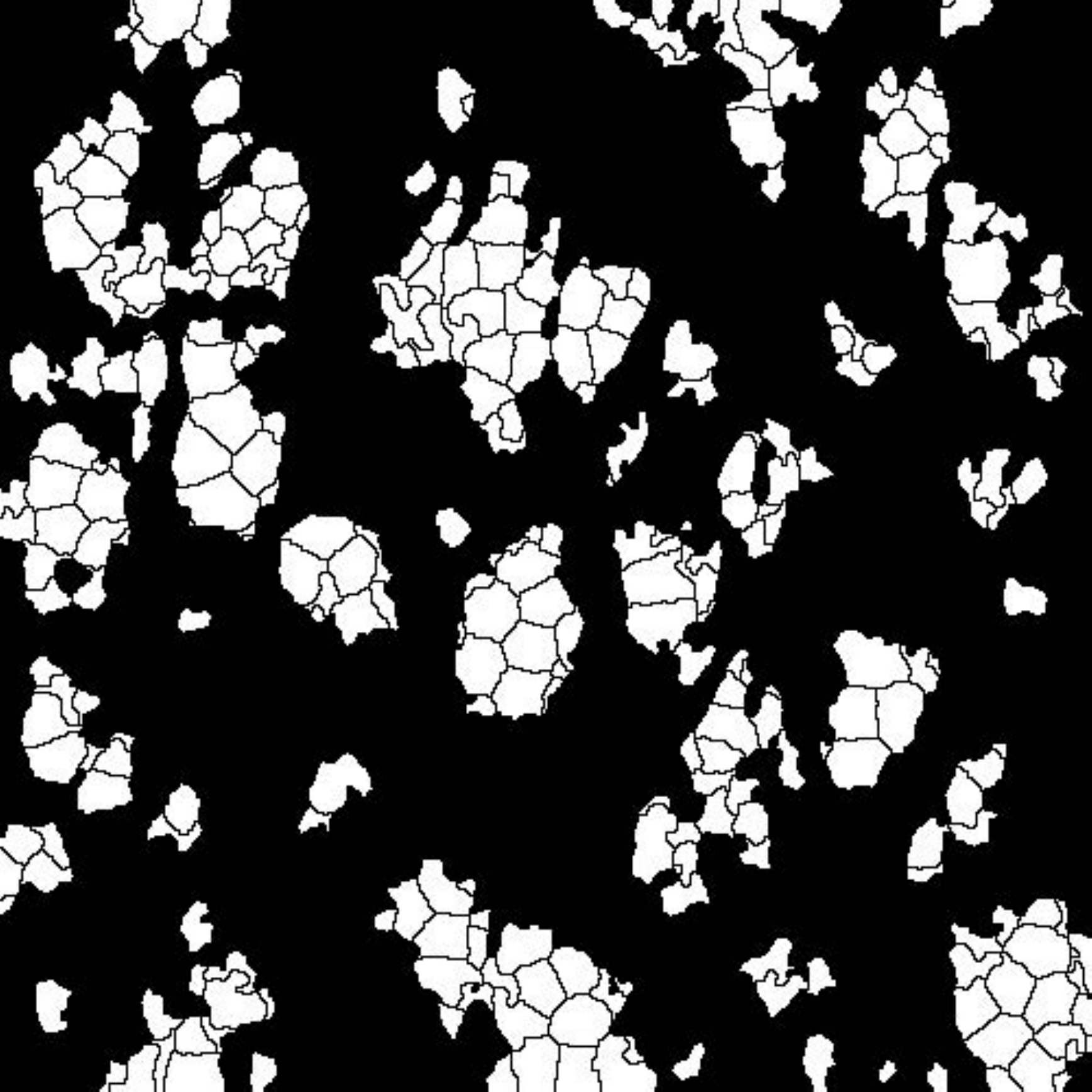}}
  \centerline{ }
\end{minipage}
\hfill
\begin{minipage}{0.16\linewidth}
  \centerline{\includegraphics[width=2.9cm]{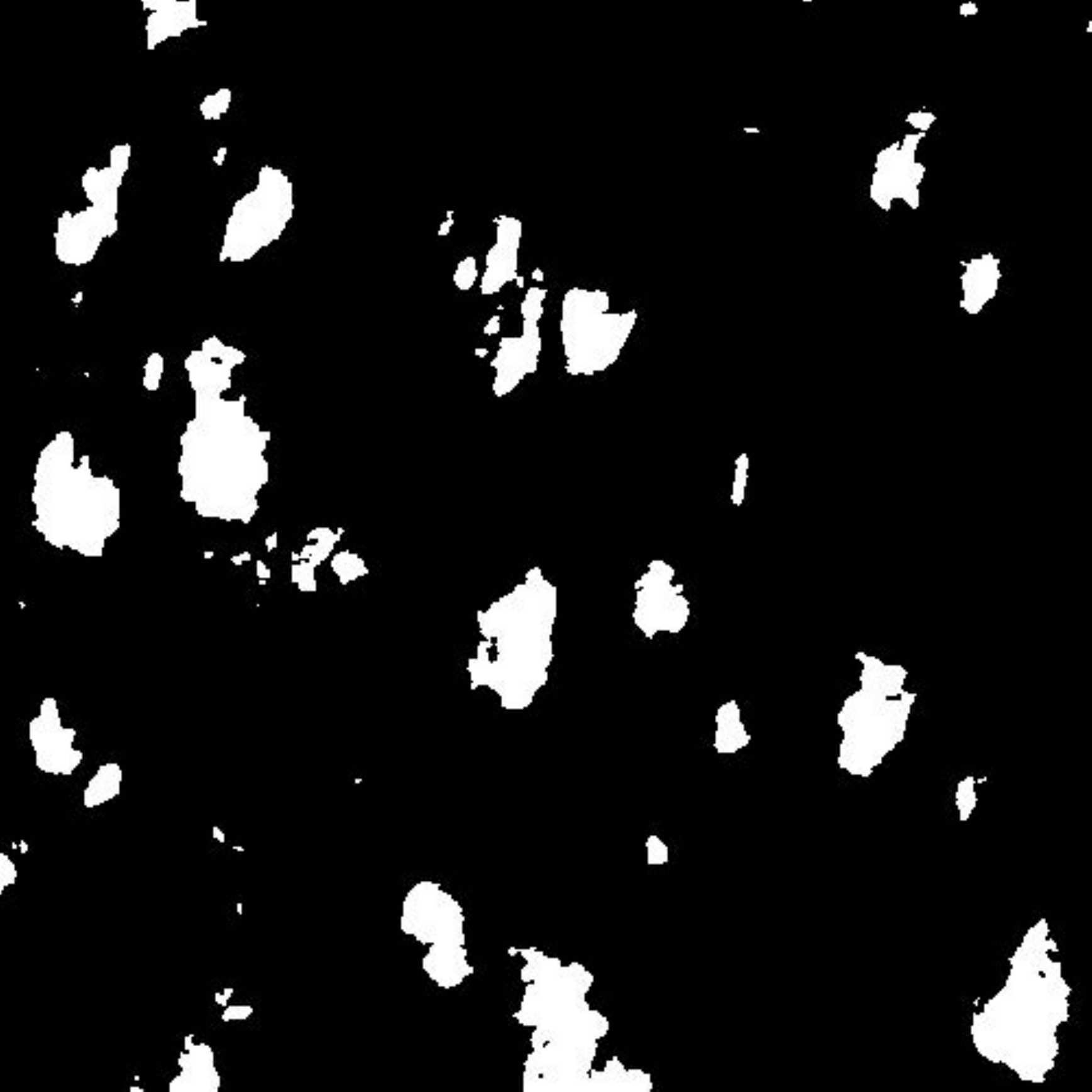}}
  \centerline{ }
\end{minipage}
\hfill
\begin{minipage}{0.16\linewidth}
  \centerline{\includegraphics[width=2.9cm]{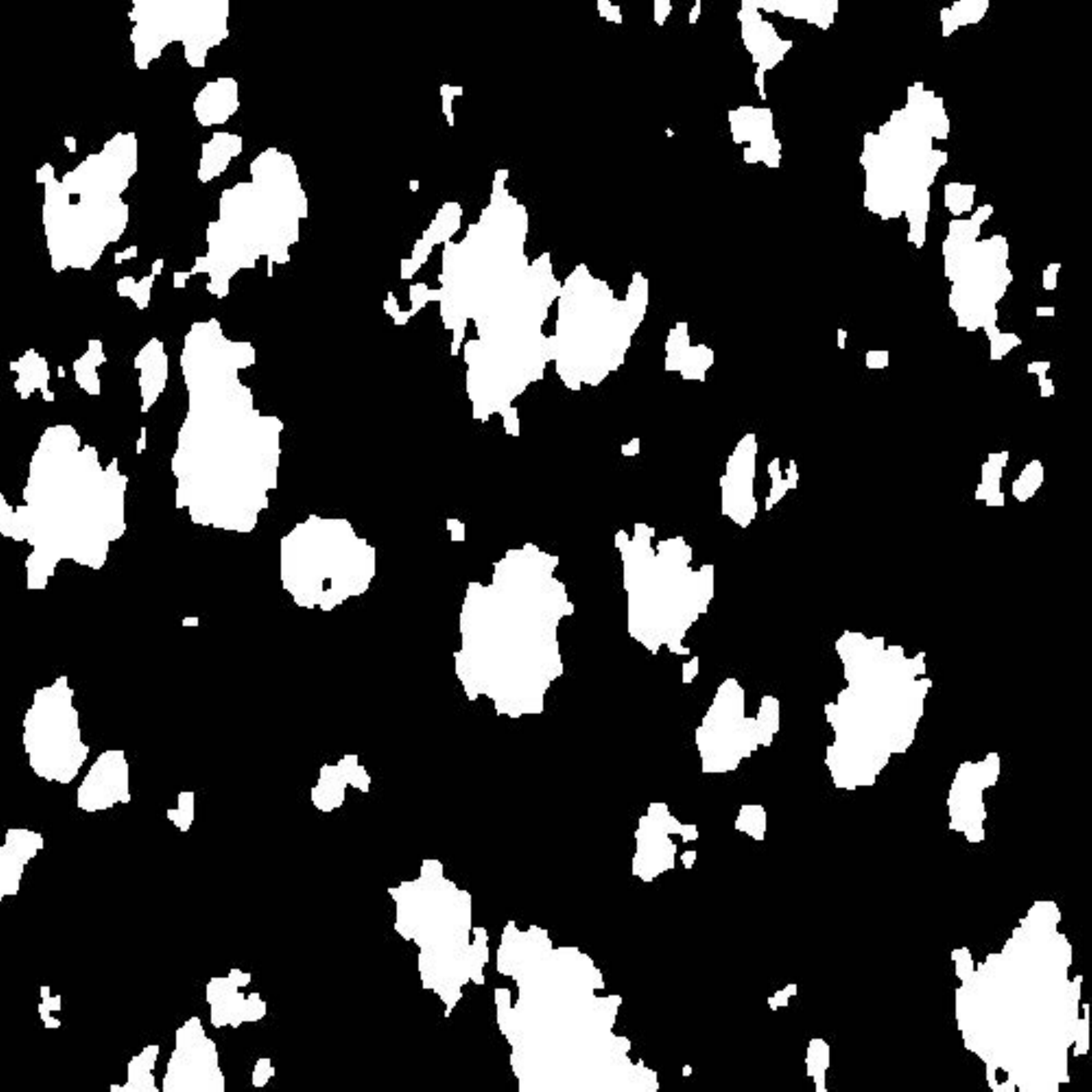}}
  \centerline{ }
\end{minipage}
\hfill
\begin{minipage}{0.16\linewidth}
  \centerline{\includegraphics[width=2.9cm]{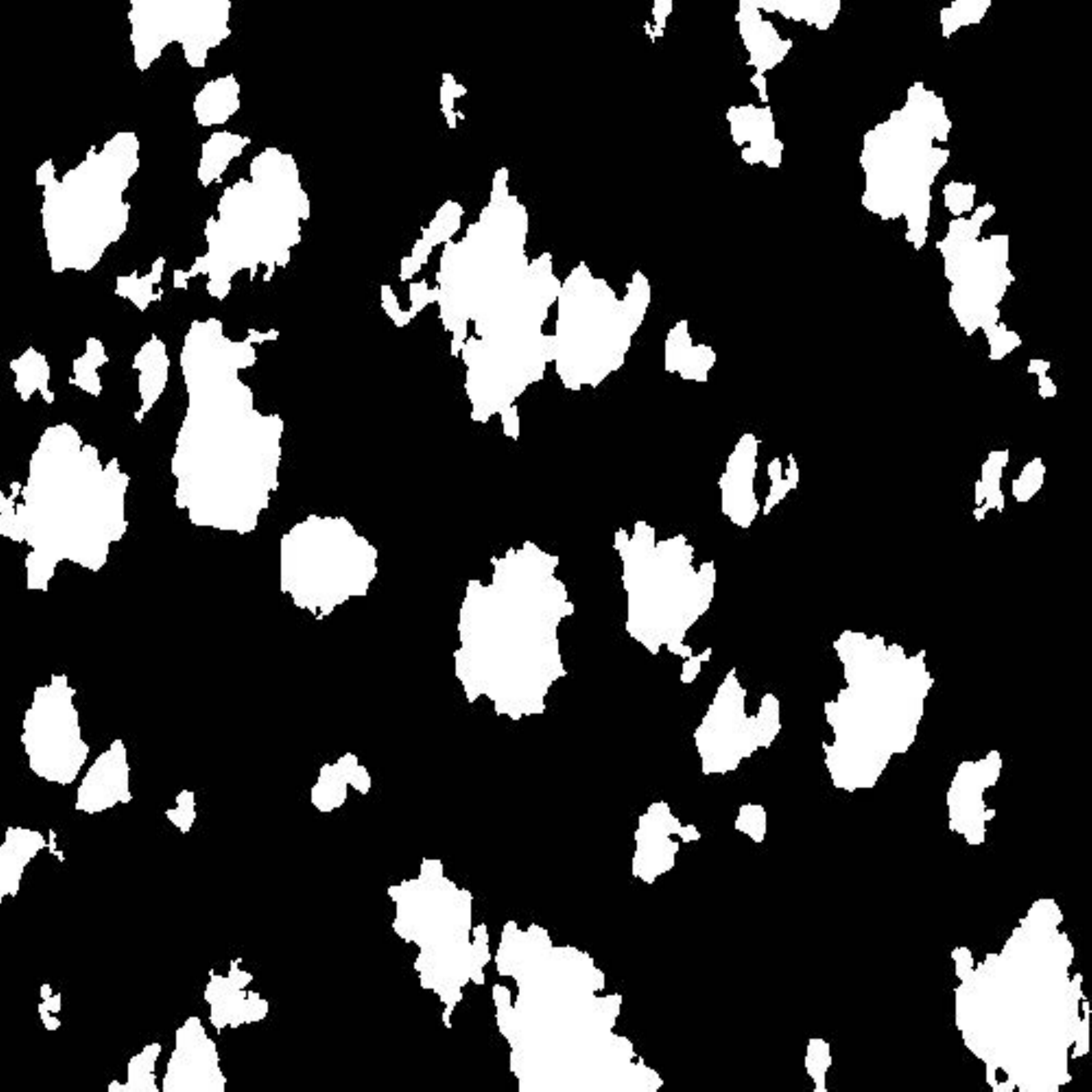}}
  \centerline{ }
\end{minipage}
\hfill
\begin{minipage}{0.16\linewidth}
  \centerline{\includegraphics[width=2.9cm]{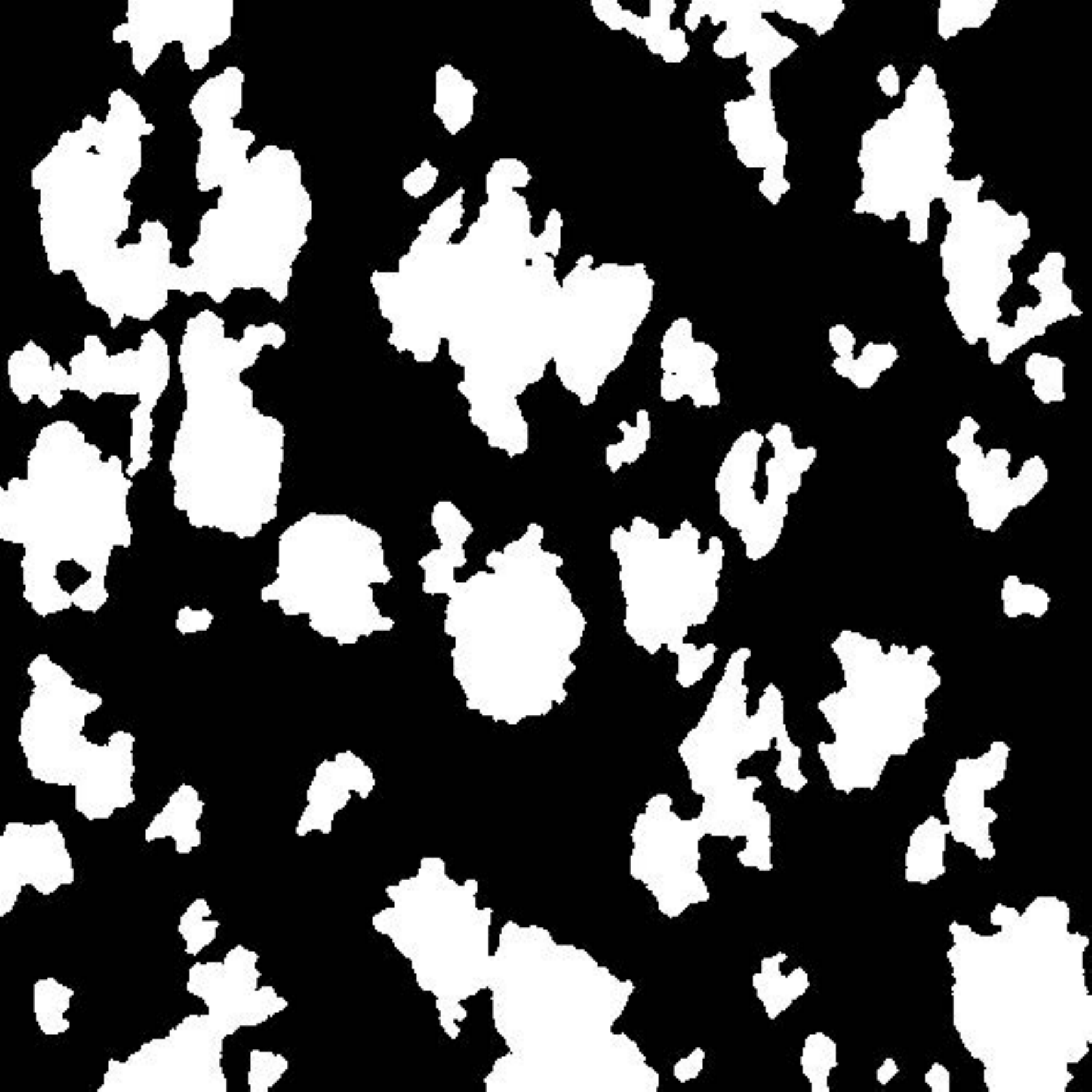}}
  \centerline{ }
\end{minipage}
\hfill
\begin{minipage}{0.16\linewidth}
  \centerline{\includegraphics[width=2.9cm]{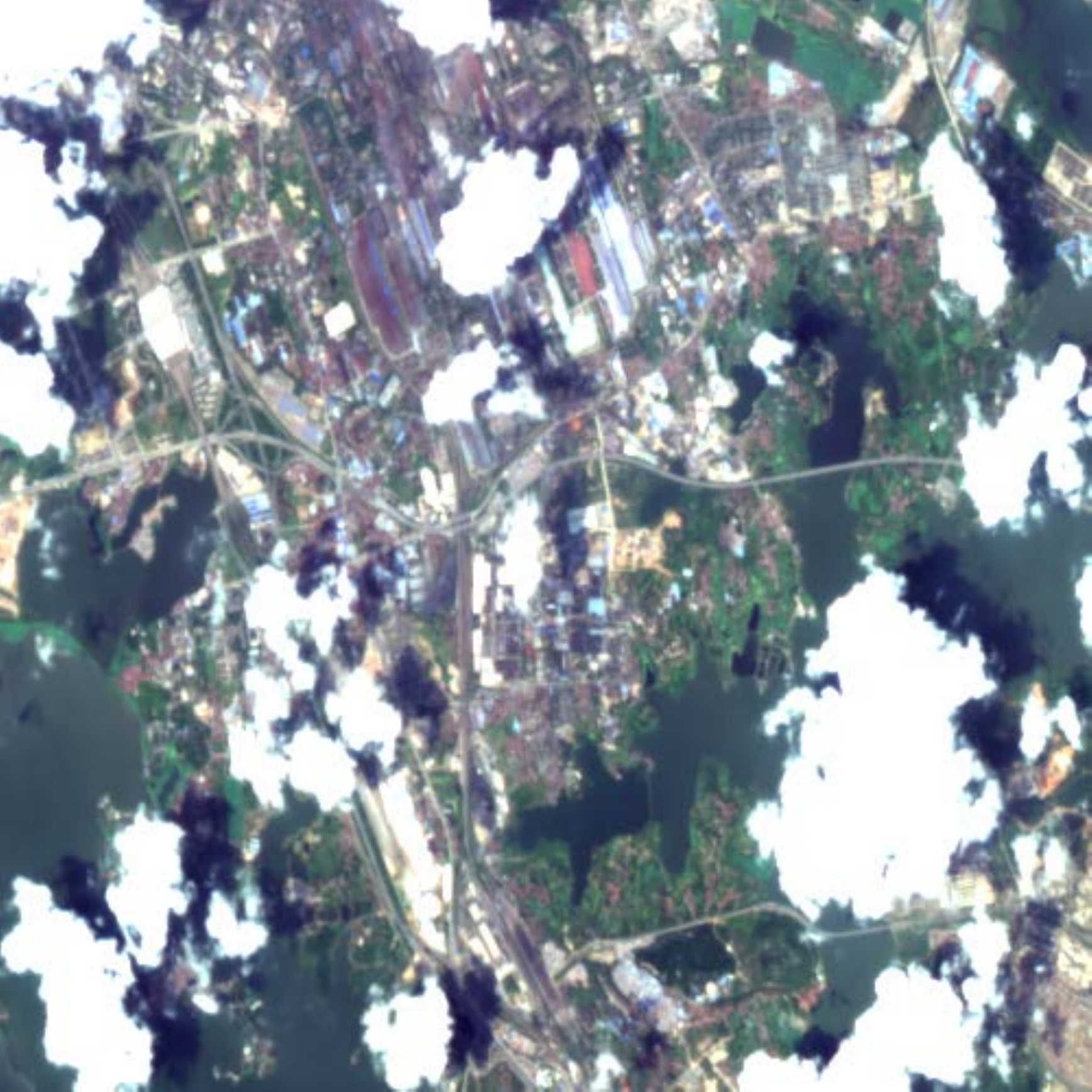}}
  \centerline{ }
\end{minipage}
\hfill
\begin{minipage}{0.16\linewidth}
  \centerline{\includegraphics[width=2.9cm]{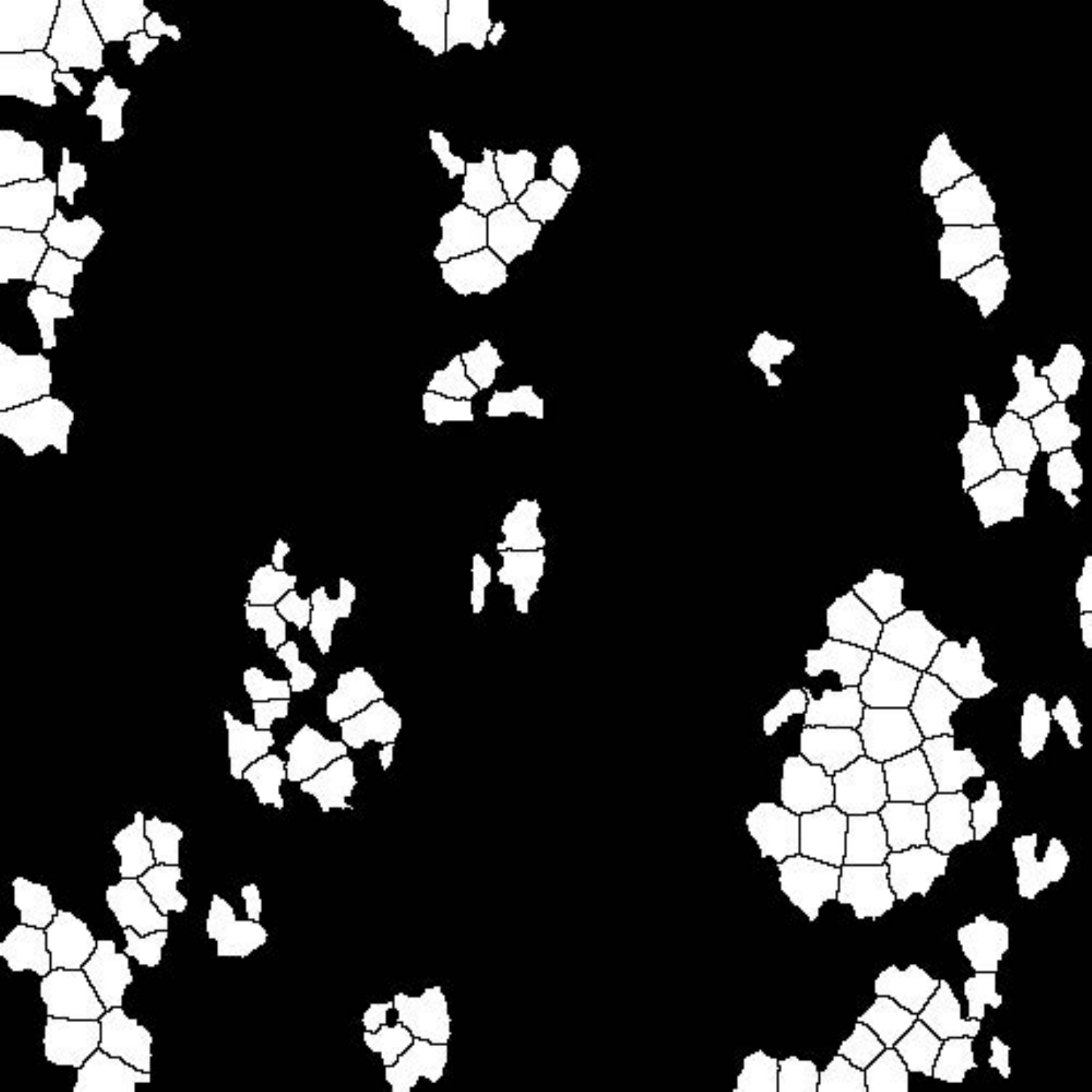}}
  \centerline{ }
\end{minipage}
\hfill
\begin{minipage}{0.16\linewidth}
  \centerline{\includegraphics[width=2.9cm]{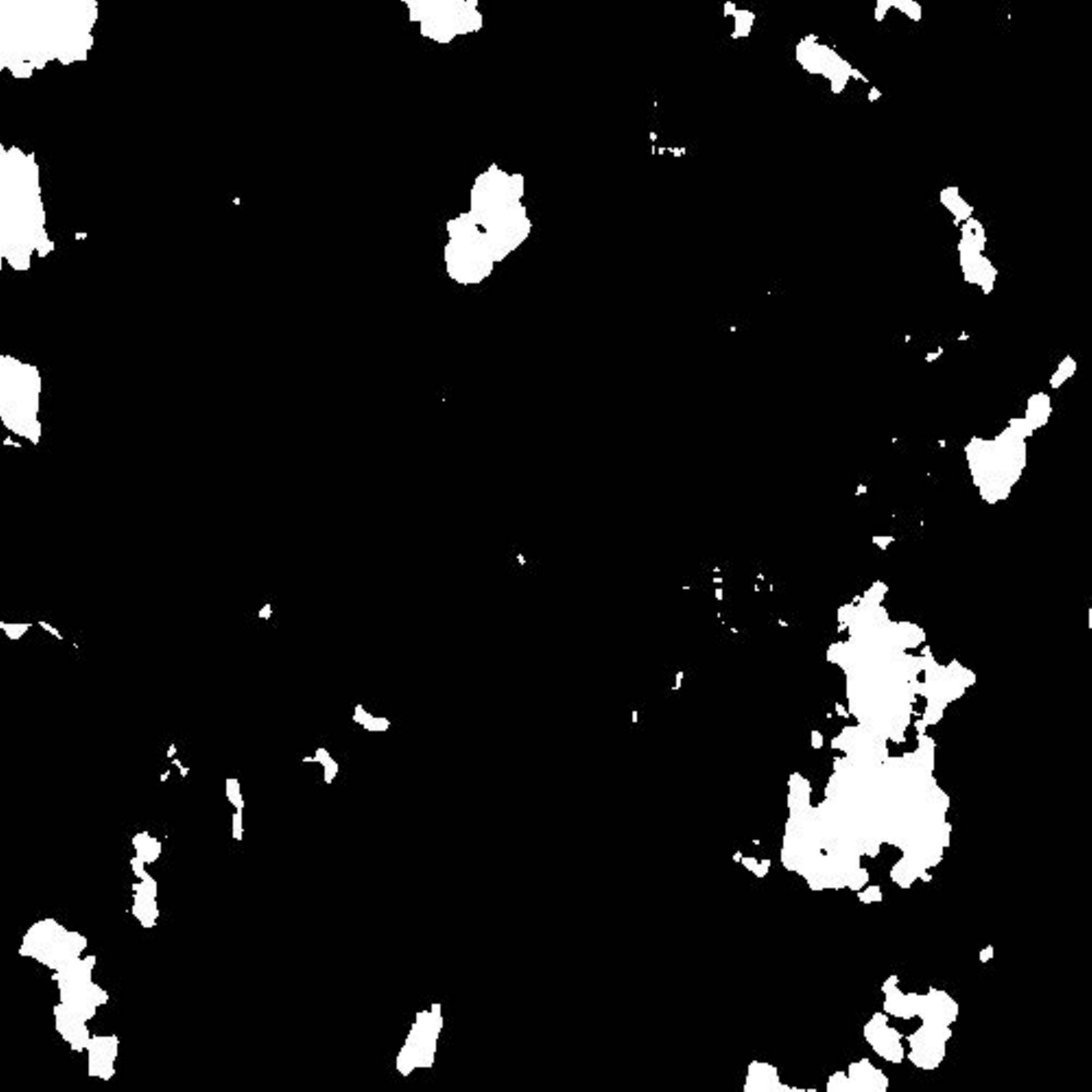}}
  \centerline{ }
\end{minipage}
\hfill
\begin{minipage}{0.16\linewidth}
  \centerline{\includegraphics[width=2.9cm]{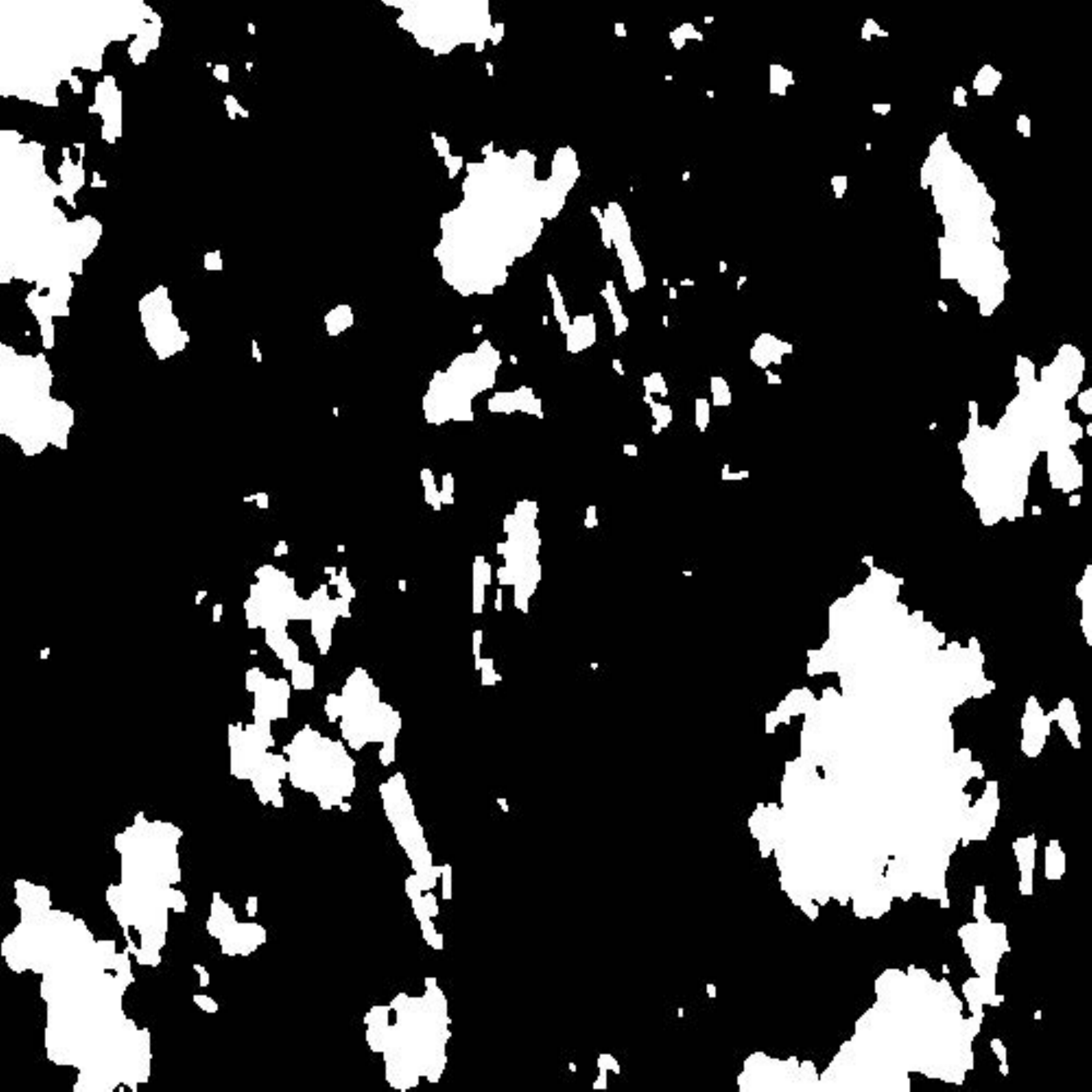}}
  \centerline{ }
\end{minipage}
\hfill
\begin{minipage}{0.16\linewidth}
  \centerline{\includegraphics[width=2.9cm]{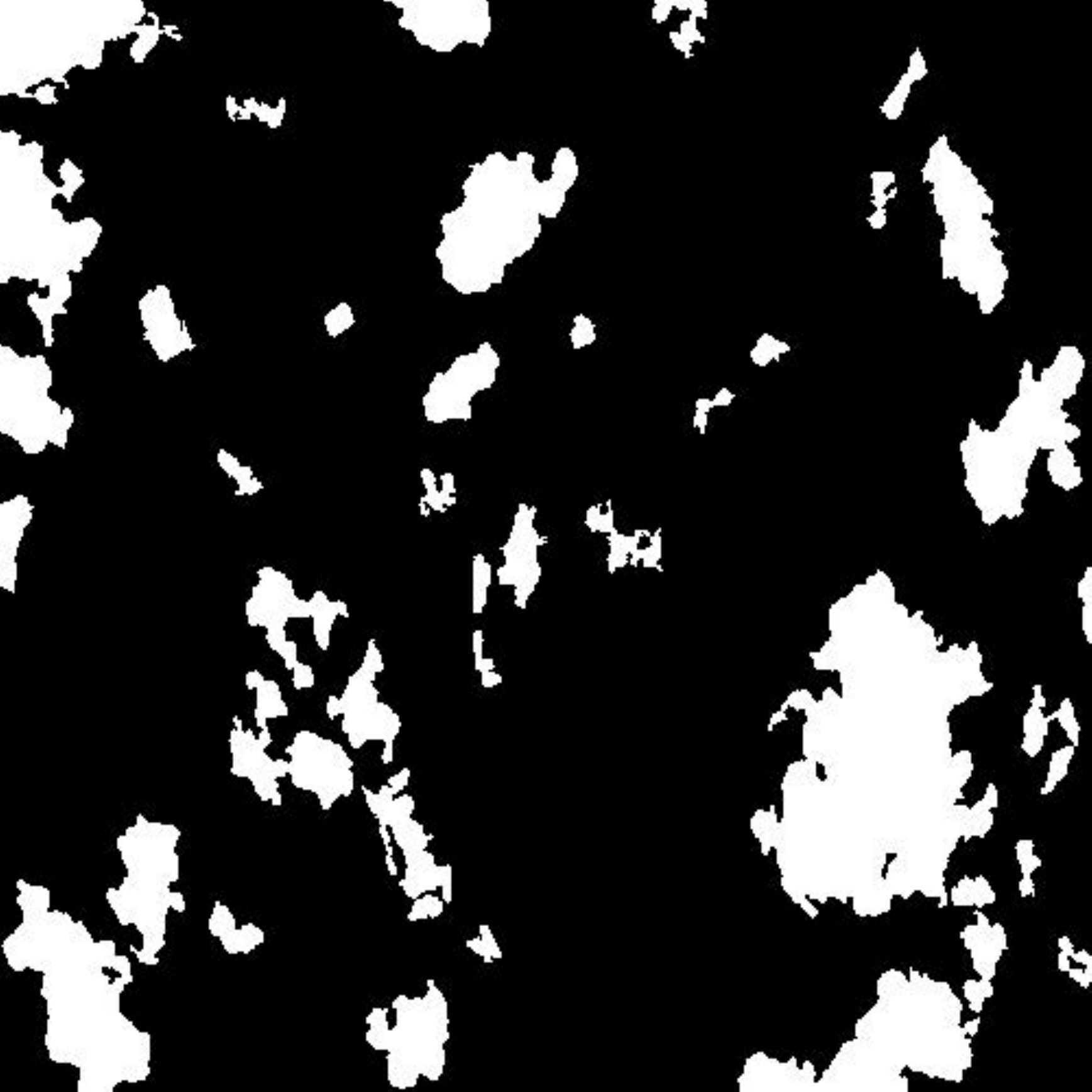}}
  \centerline{ }
\end{minipage}
\hfill
\begin{minipage}{0.16\linewidth}
  \centerline{\includegraphics[width=2.9cm]{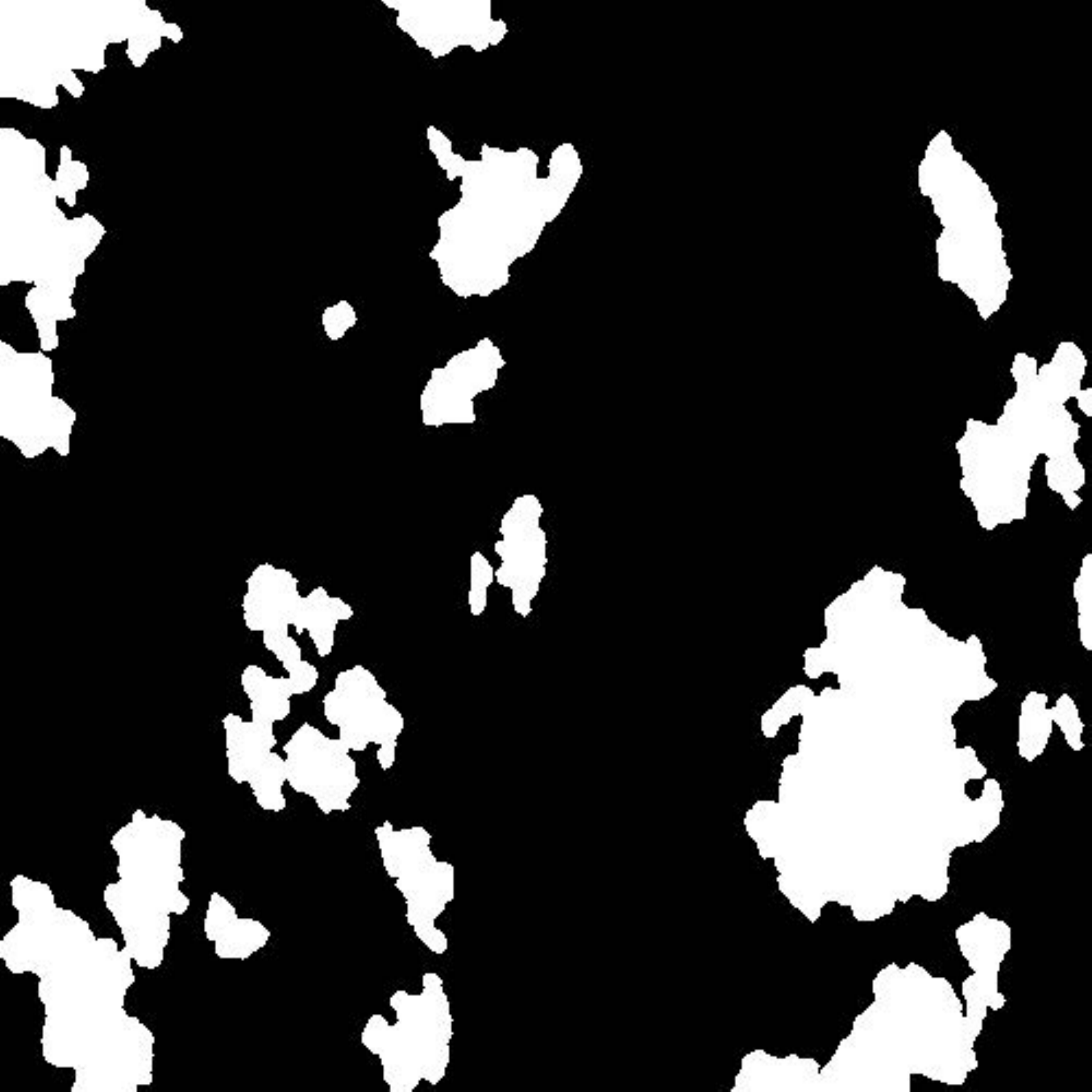}}
  \centerline{ }
\end{minipage}
\hfill
\begin{minipage}{0.16\linewidth}
  \centerline{\includegraphics[width=2.9cm]{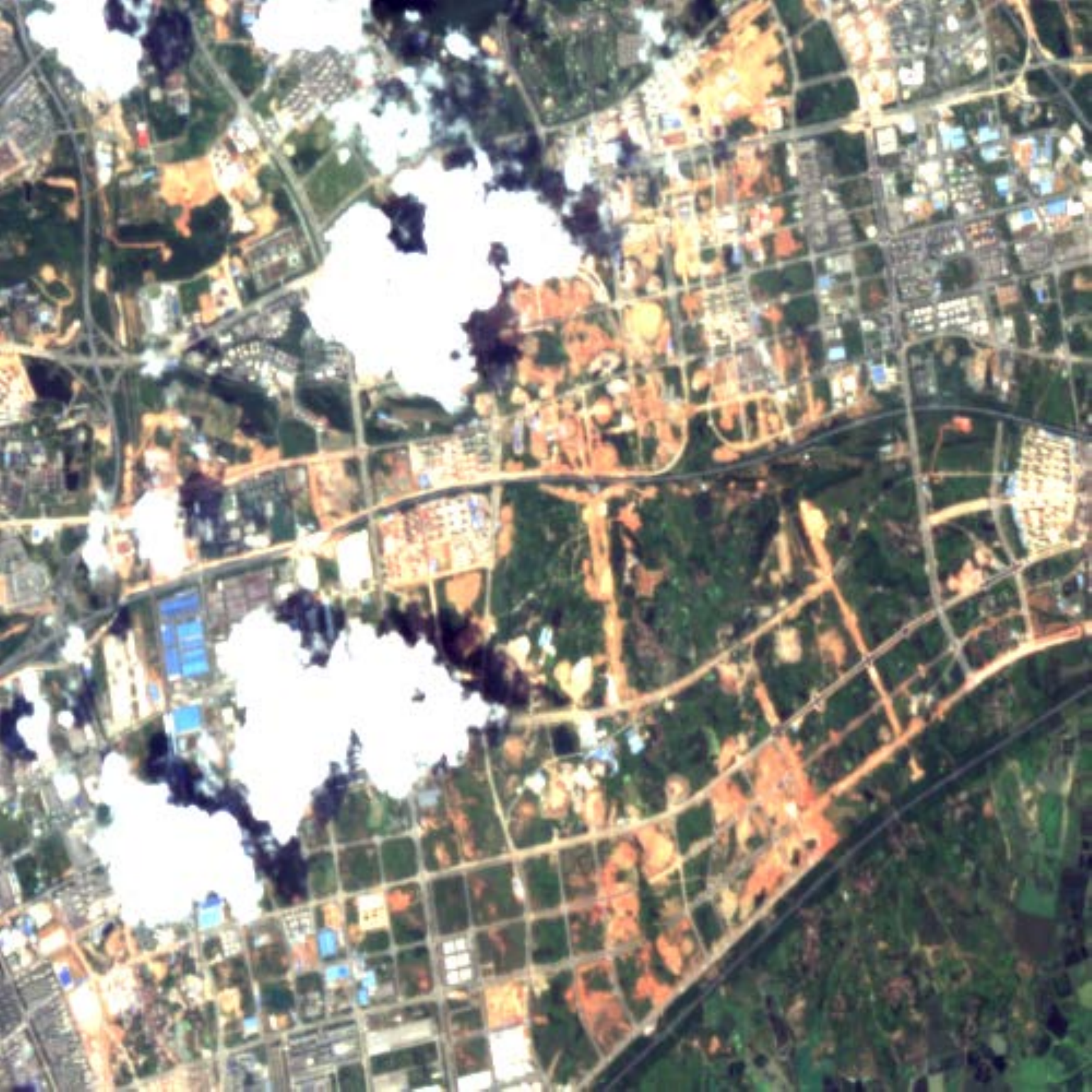}}
  \centerline{(a)}
\end{minipage}
\hfill
\begin{minipage}{0.16\linewidth}
  \centerline{\includegraphics[width=2.9cm]{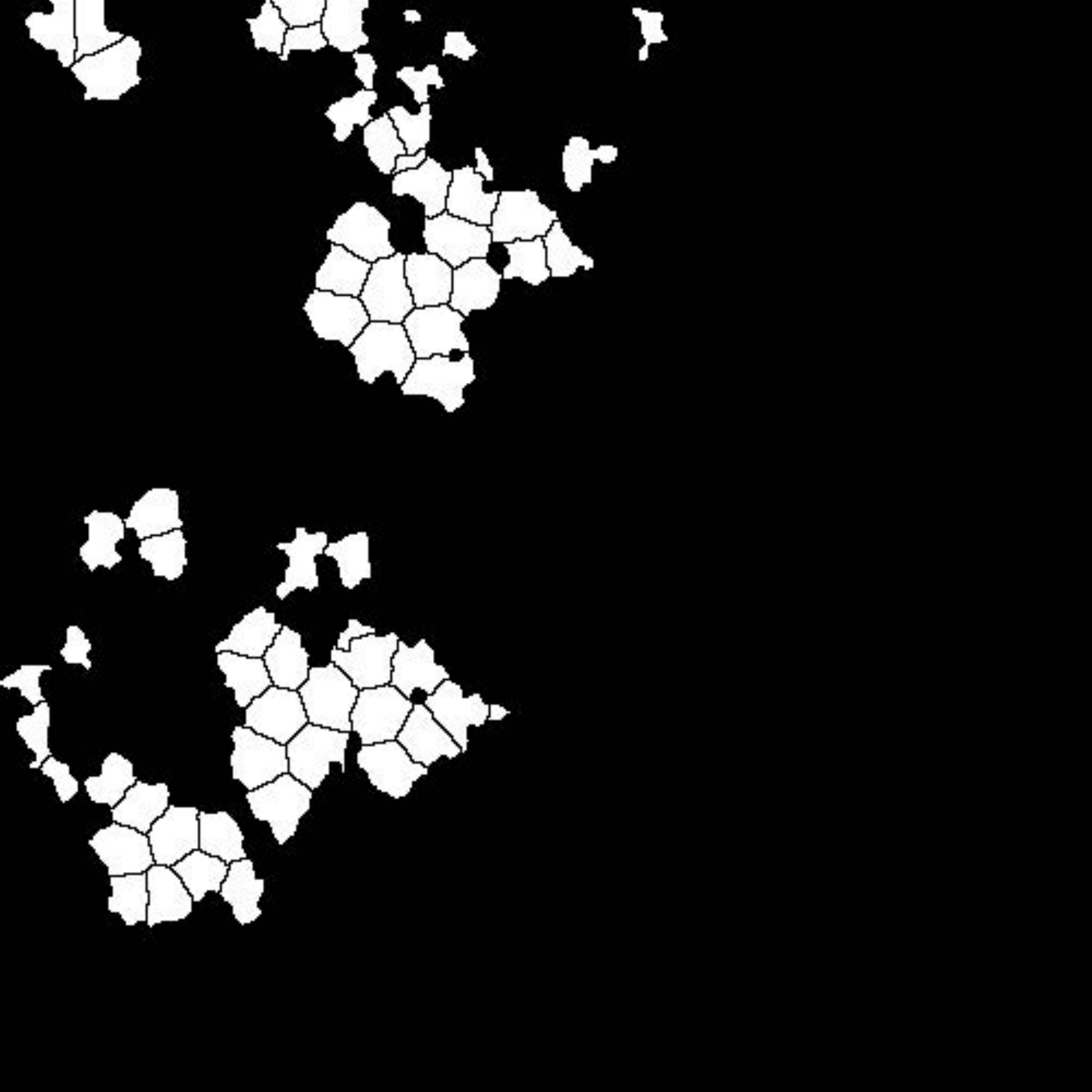}}
  \centerline{(b)}
\end{minipage}
\hfill
\begin{minipage}{0.16\linewidth}
  \centerline{\includegraphics[width=2.9cm]{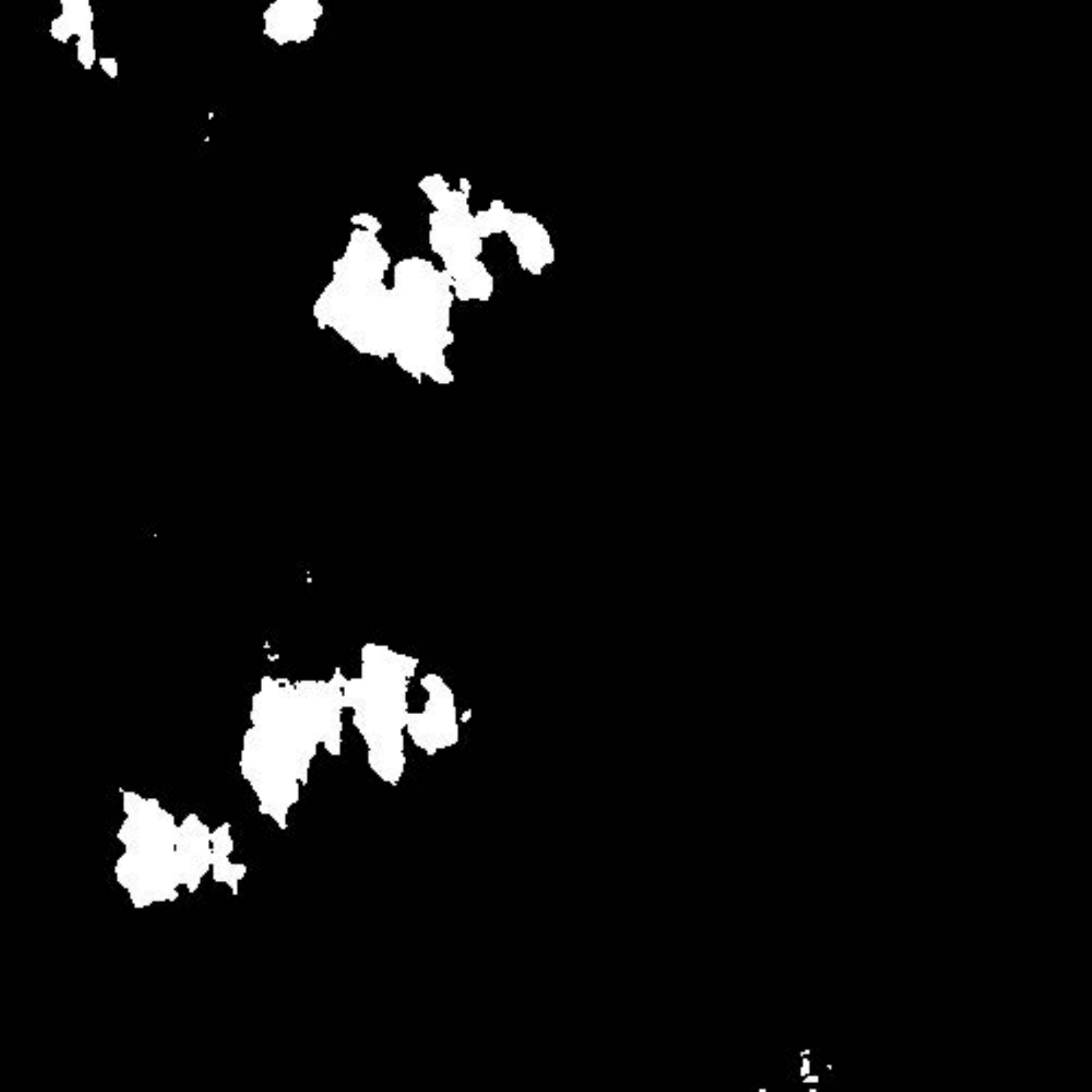}}
  \centerline{(c)}
\end{minipage}
\hfill
\begin{minipage}{0.16\linewidth}
  \centerline{\includegraphics[width=2.9cm]{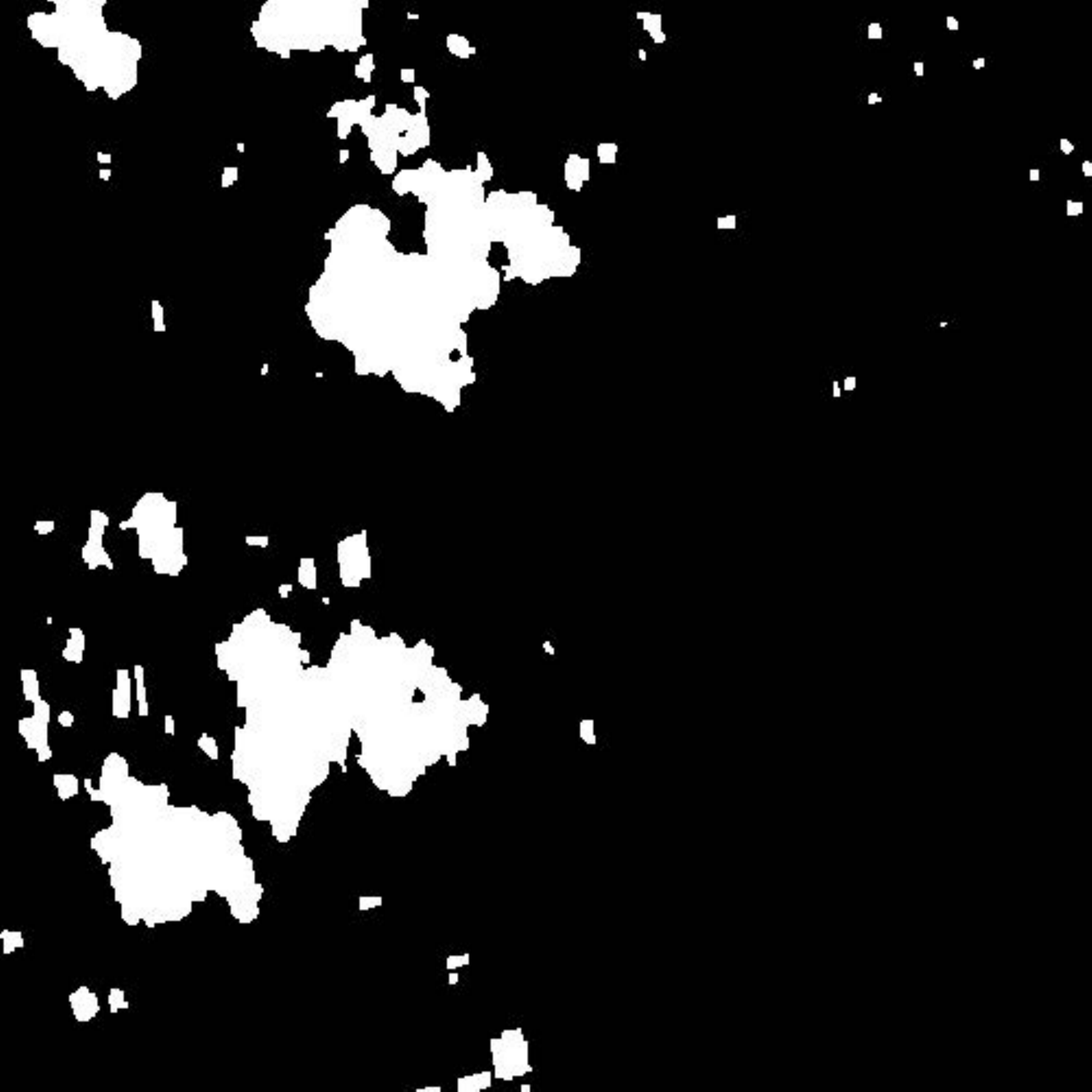}}
  \centerline{(d)}
\end{minipage}
\hfill
\begin{minipage}{0.16\linewidth}
  \centerline{\includegraphics[width=2.9cm]{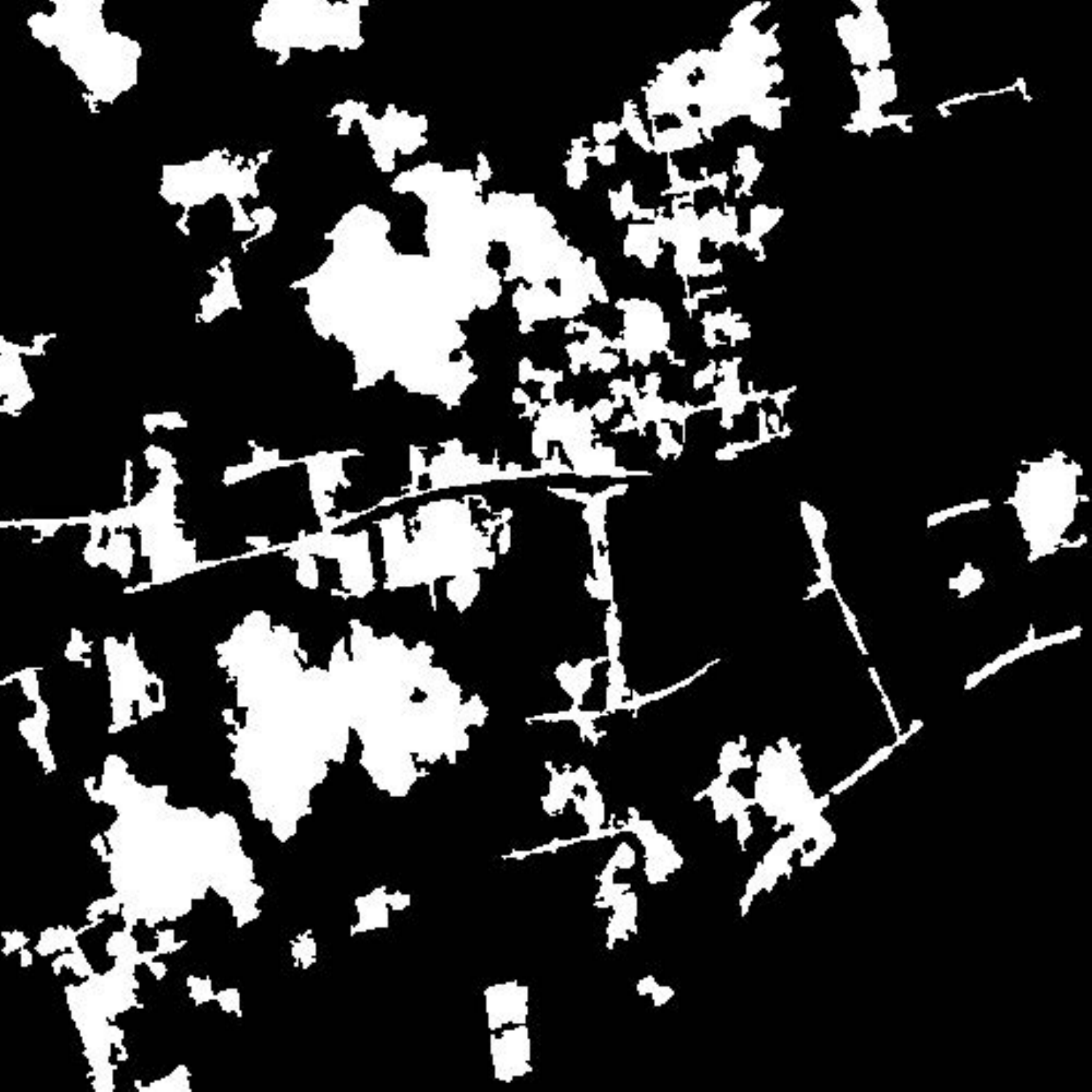}}
  \centerline{(e)}
\end{minipage}
\hfill
\begin{minipage}{0.16\linewidth}
  \centerline{\includegraphics[width=2.9cm]{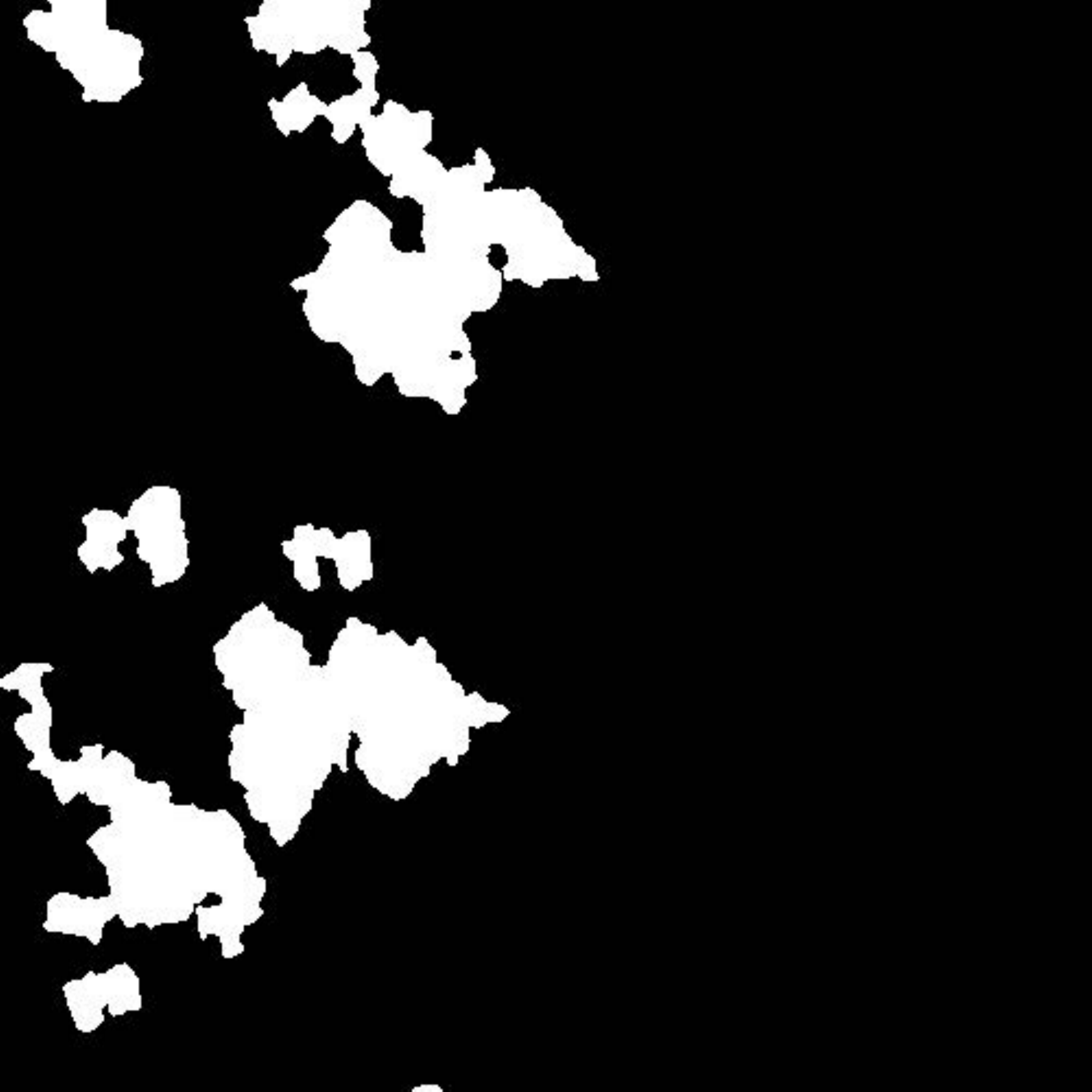}}
  \centerline{(f)}
\end{minipage}
\caption{Experimental results. (a) Original remote sensing images. (b) Groundtruth of remote sensing images. (c) Results of NIR threshold. (d) Results of GraphCut. (e) Results from\cite{b5}. (f) Results from our proposed method.}
\label{fig6}
\end{figure*}
Our proposed HFCNN is implemented with Caffe framework\cite{b15}. All remote sensing images are cropped into $500 \times 500$ first. When training HFCNN model, we use SGD optimizer. The learning rate is set to 0.001, momentum is set to 0.9, and the maximum iteration is 10000. We follow the default setting in \cite{b2} when training cascade forest. Experimental results are shown in Fig.\ref{fig6}, Fig.\ref{fig6} (a) are original GF-1 remote sensing images. Fig.\ref{fig6} (b) are groundtruth of remote sensing images for cloud detection, they are remain super-pixel level for clarity of presentation. We discussed the near-infrared threshold in section \ref{CC}, here we set the threshold to 1000, which is a statistical result in most cases. Results of NIR threshold are shown in Fig.\ref{fig6} (c), most of thick clouds are detected, but cirrus clouds are still not good enough. Fig.\ref{fig6} (d) are results from GraphCut, a classic user interactive image segmentation method. Foreground and background have to be labeled manually, clustering algorithm is then used for cloud detection. Fig.\ref{fig6} (e) are results from Zhang and Xiao \cite{b5}, in which only RGB color channels are used, cloud regions are refined progressively. Fig.\ref{fig6} (f) are results from our proposed method, gaps between super-pixels are erased to maintain consistency with other methods. From the visual aspect, our method can detect more accurate cloud regions no matter simple or complex background.

Relative precision and recall are computed in Table 1. Total super-pixel number in each remote sensing image is not fixed, so precision and recall can only be regard as relative precision and relative recall, which means they only make sense when compared with other methods in the same image. F-measure is the harmonic mean (Hmean) of recall and precision, which is always used for ranking detection tasks. Precision and recall are computed in super-pixel level. As for pixel level methods, cloud numbers are counted only if its area is more than half of a super-pixel.

Precision of NIR threshold is always very high in every remote sensing image, because it misses a large amount of cloud super-pixels, namely its recall is very low. It's interesting to find that results of GraphCut are slightly better than cloud detection method \cite{b5}. GraphCut needs manually labeled informative foreground and background, however method in \cite{b5} is automatic. Another reason might be that target remote sensing images in \cite{b5} are different from ours. Background of images in first three rows in Fig.\ref{fig6} (a) are simple but with many cirrus clouds. In this case, recalls of conventional methods are low. However, ours maintain high recall. Image in the penultimate row has few cirrus clouds but with many buildings look like clouds, which makes precision of other methods low and still ours maintain better performance. As for image in the last row, some buildings extremely resemble clouds due to illumination. GraphCut nearly takes anything in white as clouds, and method in \cite{b5} only using RGB color space even takes roads for clouds. However, our proposed method maintains better performance.

Experimental results show that our method can get excellent F-measure no matter background in remote sensing images is simple or complex, which proves the feasibility and effectiveness of our proposed method.

\begin{table}[htbp]
\centering
\caption{Quantitative Evaluation Of Accuracy And Recall For Results In Fig.\ref{fig6}}
\label{tab1}
\begin{tabular}{|c|c|c|c|c|}
\hline
Input Images                & Methods      & Precision & Recall & F-measure     \\ \hline
\multirow{4}{*}{The first}  & NIR\_1000    & 1.00         & 0.64   & 0.78          \\ \cline{2-5}
                            & GraphCut     & 1.00         & 0.80   & 0.89          \\ \cline{2-5}
                            & Zhang {[}5{]}      & 0.99      & 0.80   & 0.88          \\ \cline{2-5}
                            & Our proposed & 1.00         & 0.94   & \textbf{0.97} \\ \hline
\multirow{4}{*}{The second} & NIR\_1000    & 1.00         & 0.78   & 0.87          \\ \cline{2-5}
                            & GraphCut     & 1.00         & 0.87   & 0.93          \\ \cline{2-5}
                            & Zhang {[}5{]}      & 0.98      & 0.83   & 0.90          \\ \cline{2-5}
                            & Our proposed & 0.99      & 0.99   & \textbf{0.99} \\ \hline
\multirow{4}{*}{The third}  & NIR\_1000    & 0.90      & 0.52   & 0.66          \\ \cline{2-5}
                            & GraphCut     & 0.79      & 0.99   & 0.88          \\ \cline{2-5}
                            & Zhang {[}5{]}      & 0.81      & 0.94   & 0.87          \\ \cline{2-5}
                            & Our proposed & 0.97      & 0.94   & \textbf{0.95} \\ \hline
\multirow{4}{*}{The fourth} & NIR\_1000    & 0.99      & 0.45   & 0.62          \\ \cline{2-5}
                            & GraphCut     & 1.00         & 0.73   & 0.84          \\ \cline{2-5}
                            & Zhang {[}5{]}      & 1.00         & 0.72   & 0.83          \\ \cline{2-5}
                            & Our proposed & 0.99      & 0.98   & \textbf{0.99} \\ \hline
\multirow{4}{*}{The fifth} & NIR\_1000    & 1.00      & 0.46   & 0.63          \\ \cline{2-5}
                            & GraphCut     & 0.63         & 0.95   & 0.76          \\ \cline{2-5}
                            & Zhang {[}5{]}      & 0.33         & 0.91   & 0.49          \\ \cline{2-5}
                            & Our proposed & 0.88      & 0.91   & \textbf{0.90} \\ \hline
\end{tabular}
\end{table}

\section{Conclusions And Future Work}\label{CAFW}
This paper focuses on super-pixel level GF-1 remote sensing cloud detection task based on CNN and deep forest. Structured forests is utilized to detect edge in remote sensing images for better super-pixel segmentation. Segmented super-pixels are used to build a super-pixel level remote sensing image database. Our proposed Hierarchical Fusion CNN (HFCNN) pays more attention on low-level features since super-pixels have less semantic information. And deep forest is also implemented for cloud detection task. Both models are used to extract rich features and predict categories of super-pixels. We also propose a distance metric to refine ambiguous super-pixels such as cirrus clouds and buildings based on predictions and features from two models. Experimental results show our proposed method can achieve better results than conventional cloud detection methods.

Though our method has achieved exciting results, labeling every super-pixel in each remote sensing image is still time-consuming, an end-to-end model will be explored in the future. Shape of super-pixels are various, we will also explore using shape-adaptive super-pixels toward more precise detection result.

\section*{Acknowledgment}
This work was supported by the National Natural Science Foundation of China (No. 61572307).


\section*{References}

\begin{thebibliography}{00}
\bibitem{b1} Xie F, Shi M, Shi Z, et al. Multilevel Cloud Detection in Remote Sensing Images Based on Deep Learning[J]. IEEE Journal of Selected Topics in Applied Earth Observations and Remote Sensing, 2017, 10(8): 3631-3640.
\bibitem{b2} Zhou Z H, Feng J. Deep forest: Towards an alternative to deep neural networks[J]. arXiv preprint arXiv:1702.08835, 2017.
\bibitem{b3} Doll¨¢r P, Zitnick C L. Fast Edge Detection Using Structured Forests.[J]. IEEE Transactions on Pattern Analysis \& Machine Intelligence, 2015, 37(8):1558-70.
\bibitem{b4} Jedlovec G J, Haines S L, Lafontaine F J. Spatial and Temporal Varying Thresholds for Cloud Detection in GOES Imagery[J]. IEEE Transactions on Geoscience \& Remote Sensing, 2008, 46(6):1705-1717.
\bibitem{b5} Zhang Q, Xiao C. Cloud Detection of RGB Color Aerial Photographs by Progressive Refinement Scheme[J]. IEEE Transactions on Geoscience \& Remote Sensing, 2014, 52(11):7264-7275.
\bibitem{b6} Visa A, Valkealahti K, Simula O. Cloud detection based on texture segmentation by neural network methods[C]// IEEE International Joint Conference on Neural Networks. IEEE Xplore, 1991:1001-1006 vol.2.p
\bibitem{b7} Hu X, Wang Y, Shan J. Automatic Recognition of Cloud Images by Using Visual Saliency Features[J]. IEEE Geoscience \& Remote Sensing Letters, 2015, 12(8):1760-1764.
\bibitem{b8} Tan K, Zhang Y, Tong X. Cloud Extraction from Chinese High Resolution Satellite Imagery by Probabilistic Latent Semantic Analysis and Object-Based Machine Learning[J]. Remote Sensing, 2016, 8(11):963.
\bibitem{b9} Yuan Y, Hu X. Bag-of-Words and Object-Based Classification for Cloud Extraction From Satellite Imagery[J]. IEEE Journal of Selected Topics in Applied Earth Observations \& Remote Sensing, 2015, 8(8):4197-4205.
\bibitem{b10} Perfetto Q. Classifying Handwritten Digits Using Random Forests[J]. 2017.
\bibitem{b11} Zhang C, Yan J, Li C, et al. Contour detection via stacking random forest learning[J]. Neurocomputing, 2018, 275: 2702-2715.
\bibitem{b12} Shen Y, Lu H, Jia J. Classification of Motor Imagery EEG Signals with Deep Learning Models[C]//International Conference on Intelligent Science and Big Data Engineering. Springer, Cham, 2017: 181-190.
\bibitem{b13} Krizhevsky A, Hinton G. Learning multiple layers of features from tiny images[J]. 2009.
\bibitem{b14} Lin T Y, Doll¨¢r P, Girshick R, et al. Feature pyramid networks for object detection[C]//CVPR. 2017, 1(2): 4.
\bibitem{b15} Jia, Yangqing, Shelhamer, et al. Caffe: Convolutional Architecture for Fast Feature Embedding[J]. 2014:675-678.


\end{thebibliography}
\end{document}